\documentclass[letterpaper]{article}
\usepackage{uai2019}
\usepackage[margin=1in]{geometry}

\usepackage{times}

\usepackage{amsmath,amsfonts,bm}

\def\eqref#1{equation~\ref{#1}}
\def\1{\bm{1}}

\DeclareMathAlphabet{\mathsfit}{\encodingdefault}{\sfdefault}{m}{sl}
\SetMathAlphabet{\mathsfit}{bold}{\encodingdefault}{\sfdefault}{bx}{n}

\newcommand{\E}{\mathbb{E}}

\newcommand{\KL}{D_{\mathrm{KL}}}

\DeclareMathOperator*{\argmax}{arg\,max}

\usepackage[numbers,sort&compress]{natbib}
\usepackage{hyperref}
\usepackage{url}
\usepackage{paralist}
\usepackage[acronym,smallcaps,nowarn,section,nogroupskip,nonumberlist]{glossaries}
\usepackage[nameinlink,capitalise]{cleveref}
\usepackage{amssymb}
\usepackage{mathtools}
\usepackage{tikz}
\usepackage{etoolbox}
\usepackage{graphicx}
\usepackage{wrapfig}
\usepackage[font=small]{caption}
\usepackage{subcaption}
\usepackage{booktabs}
\usepackage{titlesec}
\usepackage[textsize=footnotesize]{todonotes}
\usepackage{balance}
\usepackage[ruled,vlined]{algorithm2e}

\definecolor{red}{HTML}{E41A1C}
\definecolor{orange}{HTML}{FF7F00}
\definecolor{yellow}{HTML}{FFC020}
\definecolor{green}{HTML}{4DAF4A}
\definecolor{blue}{HTML}{377EB8}
\definecolor{purple}{HTML}{984EA3}
\usetikzlibrary{calc,backgrounds,positioning,bayesnet,arrows.meta,patterns,fit,trees}
\presetkeys{todonotes}{%
  backgroundcolor=blue!10!white,
  linecolor=blue!10!white,
  bordercolor=blue!10!white
}{}
\Crefname{algocf}{Algorithm}{Algorithms}
\crefname{algorithm}{Algorithm}{Algorithms}
\crefname{figure}{Figure}{Figure}
\crefname{table}{Table}{Table}
\crefname{section}{\S}{\S\S}
\Crefname{section}{\S}{\S\S}
\crefformat{equation}{(#2#1#3)}
\makeglossaries
\glsdisablehyper{}
\newacronym{SCFM}{scfm}{stochastic control-flow model}
\newacronym{WS}{ws}{wake-sleep}
\newacronym{BWS}{bws}{basic wake-sleep}
\newacronym{RWS}{rws}{reweighted wake-sleep}
\newacronym{ELBO}{elbo}{evidence lower bound}
\newacronym{VAE}{vae}{variational autoencoder}
\newacronym{IWAE}{iwae}{importance weighted autoencoder}
\newacronym{KL}{kl}{Kullback-Leibler}
\newacronym{SGD}{sgd}{stochastic gradient descent}
\newacronym{VIMCO}{vimco}{variational inference for Monte Carlo objectives}
\newacronym{WW}{ww}{wake-wake}
\newacronym{WWS}{wws}{wake-wake-sleep}
\newacronym{AIR}{air}{Attend, Infer, Repeat}
\newacronym{ESS}{ess}{effective sample size}
\newacronym{REINFORCE}{reinforce}{Reinforce gradient estimator}
\newacronym{IS}{is}{importance sampling}
\newacronym{GMM}{gmm}{Gaussian mixture model}
\newacronym{MNIST}{mnist}{hand-written digit dataset}
\newacronym{RELAX}{relax}{RELAX gradient estimator}
\newacronym{REBAR}{rebar}{REBAR gradient estimator}
\newacronym{PMF}{pmf}{probability mass function}
\newacronym{MLP}{mlp}{multilayer perceptron}
\newacronym{RNN}{rnn}{recurrent neural network}
\newacronym{PCFG}{pcfg}{probabilistic context free grammar}
\newacronym{ADAM}{adam}{ADAM}
\glsunset{ADAM}

\makeatletter
\renewcommand\paragraph{\@startsection{paragraph}{4}{\z@}%
 {0ex \@plus1ex \@minus.2ex}%
 {-1em}%
 {\normalfont\normalsize\bfseries}}
\makeatother

\newcommand{\circled}[2][]{%
  \tikz[baseline=(char.base)]{%
    \node[shape = circle, draw, inner sep = 1pt,scale=0.75]
    (char) {\phantom{\ifblank{#1}{#2}{#1}}};%
    \node at (char.center) {\makebox[0pt][c]{\scriptsize #2}};}}
\robustify{\circled}

\newcommand{\given}{\lvert}
\DeclareMathOperator{\ELBO}{\acrshort{ELBO}}
\DeclareMathOperator{\std}{\mathrm{std}}

\title{Revisiting Reweighted Wake-Sleep \\[0.5ex]
for Models with Stochastic Control Flow}

\author{
  Tuan~Anh Le$^1$\thanks{\quad Equal contribution.} \quad Adam R. Kosiorek$^{1, 2}$\footnotemark[1] \quad N. Siddharth$^1$\quad Yee~Whye Teh$^2$\quad Frank Wood$^3$\\
  $^1$ Department of Engineering Science, University of Oxford \\
  $^2$ Department of Statistics, University of Oxford \\
  $^3$ Department of Computer Science, University of British Columbia
}

\begin{document}

\maketitle

\begin{abstract}
  \Glspl{SCFM} are a class of generative models that involve branching on choices from discrete random variables.
  Amortized gradient-based learning of \glspl{SCFM} is challenging as most approaches targeting discrete variables rely on their continuous relaxations---which can be intractable in \glspl{SCFM}, as branching on relaxations requires evaluating \emph{all} (exponentially many) branching paths.
  Tractable alternatives mainly combine \acrshort{REINFORCE} with complex control-variate schemes to improve the variance of na\"ive estimators.
  Here, we revisit the \gls{RWS}~\citep{bornschein2015reweighted} algorithm, and through extensive evaluations, show that it outperforms current state-of-the-art methods in learning \glspl{SCFM}.
  Further, in contrast to the \acrlong{IWAE}, we observe that \gls{RWS} learns better models \emph{and} inference networks with increasing numbers of particles.
  Our results suggest that \gls{RWS} is a competitive, often preferable, alternative for learning \glspl{SCFM}.
\end{abstract}

\glsresetall

\vspace*{-1\baselineskip}
\section{INTRODUCTION}
\label{sec:introduction}

\Glspl{SCFM} describe generative models that employ branching (i.e., the use of {\small \texttt{if} / \texttt{else} / \texttt{cond}} statements) on choices from discrete random variables.
Recent years have seen such models gain relevance, particularly in the domain of deep probabilistic programming~\citep[Ch. 7]{siddharth2017learning,bingham2019pyro,tran2017deep,vandemeent2018intro}, which allows combining neural networks with generative models expressing arbitrarily complex control flow.
\Glspl{SCFM} are encountered in a wide variety of tasks including tracking and prediction \citep{neiswanger2014dependent,kosiorek2018sequential}, clustering \citep{rasmussen2000infinite}, topic modeling \citep{blei2003latent}, model structure learning \citep{adams2010learning}, counting \citep{eslami2016attend}, attention \citep{xu2015show}, differentiable data structures \citep{graves2014neural,graves2016hybrid,grefenstette2015learning}, speech \& language modeling \citep{juang1991hidden,chater2006probabilistic}, and concept learning \citep{kemp2006learning,lake2018emergence}.

While a variety of approaches for amortized gradient-based learning (targeting model \gls{ELBO}) exist for models using discrete random variables, the majority rely on continuous relaxations of the discrete variables~\citep[e.g.][]{rolfe2016dvae,vahdat2018dvaepp,vahdat2018dvaehash,oord2017neural,maddison2017concrete,jang2017categorical}, enabling gradient computation through reparameterization~\citep{kingma2014auto,rezende2014stochastic}.
The models used by these approaches typically do not involve any control flow on discrete random choices, instead choosing to feed choices from their continuous relaxations directly into a neural network, thereby facilitating the required learning.

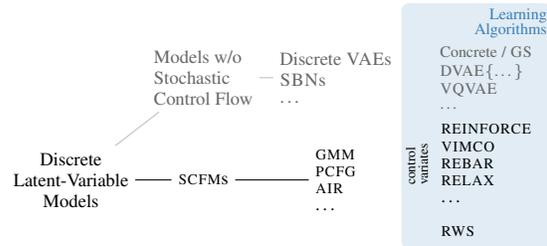
\begin{figure}[t]
  \centering
  \scriptsize
  \def\figuresize{5mm}
  \begin{tikzpicture}[%
    scale=0.8,
    txt/.style={text centered},
    infs/.style={txt,align=left,font=\tiny},
    bgb/.style={rounded corners=1mm,minimum height=1.5cm,minimum width=2cm},
    level 1/.style={sibling distance=1.7cm, level distance=2.2cm},
    level 2/.style={sibling distance=1cm, level distance=2.2cm},
    level 3/.style={level distance=2.5cm},
    grow=right,
    ]
    \node[txt,align=center] {Discrete\\Latent-Variable\\Models}
    child[missing]              % dummy child for spacing
    child {                     % SCFMs
      node[txt] {\acrshortpl{SCFM}}
      child {
        node[txt,align=left,font=\scriptsize] (eg2) {
          \acrshort{GMM}\\[-0.5ex]
          \acrshort{PCFG}\\[-0.5ex]
          \acrshort{AIR}\\[-0.5ex]
          \ldots
        }
        child {
          node[infs,font=\scriptsize] (cvs) {
            \acrshort{REINFORCE}\\[-0.5ex]
            \acrshort{VIMCO}\\[-0.5ex]
            \acrshort{REBAR}\\[-0.5ex]
            \acrshort{RELAX}\\[-0.5ex]
            \ldots\\[1.5ex]
            \acrshort{RWS}
          }
          edge from parent[draw=none]
          node[above,font=\tiny,scale=0.85,align=left,
               rotate=90,xshift=2.5mm,yshift=-5.5mm]{control\\[-0.5ex]variates}
        }
      }
    }
    child {                     % non-SCFMs
      node[txt,align=left,text=gray!80!black] {Models w/o\\Stochastic\\Control Flow}
      child {
        node[txt,text=gray!80!black,align=left] (eg1) {
          Discrete \textsc{VAE}s\\
          \textsc{SBN}s\\[-0.5ex]
          \ldots
        }
        child {
          node[infs,text=gray!80!black] (others) {
            Concrete\;/\;\textsc{GS}\\
            \textsc{DVAE}\{\ldots\}\\
            \textsc{VQVAE}\\[-0.5ex]
            \ldots
          }
         edge from parent[draw=none]
        }
      }
      edge from parent[draw=gray!50]
    };
    \scoped[on background layer]
    \node[bgb, fill=blue!15, fit={($(others.north west)+(-3mm,5mm)$)($(cvs.south east)$)},
    label={[xshift=4.5mm,yshift=-5.5mm,align=right,font=\tiny,text=blue]above:
            Learning\\[-0.5mm]Algorithms}
    ] {};
  \end{tikzpicture}
  \vspace*{-0.5\baselineskip}
  \caption{An overview of learning algorithms for discrete latent-variable models, with focus on \glspl{SCFM}.}
  \label{fig:scfm}
  \vspace*{-2.7\baselineskip}
\end{figure}

\begin{figure}[t]
  \centering\scriptsize
  \begin{tikzpicture}[%
    scale=0.9,
    >=latex,
    txt/.style={text centered,inner sep=1pt},
    nd/.style={txt,label={above:\includegraphics[width=1.2em]{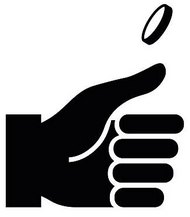}}},
    lf/.style={txt,shape=circle,draw,inner sep=2pt, scale=0.8},
    ex/.style={txt,label={right:\tikz{\draw[->,>=latex,thick,red,densely dashed] (0,0)--(0.7,0);}}},
    rr/.style={thick,transform canvas={xshift=-0.5mm,yshift=-1.1mm}},
    rl/.style={thick,transform canvas={xshift=-0.5mm,yshift=1.1mm}},
    cr/.style={thick, transform canvas={xshift=0.8mm,yshift=0.8mm}},
    cl/.style={thick, transform canvas={xshift=0.8mm,yshift=-0.8mm}},
    level 1/.style={level distance=2.2cm, sibling distance=1.9cm},
    level 2/.style={level distance=2.2cm, sibling distance=0.8cm},
    grow'=right,
    ]
    %% base tree
    \node[nd] (c1) {\(\mathtt{if}(c_1)\)}
    child {
      node[nd] (c2) {\(\mathtt{if}(c_2)\)}
      child {node[ex] (e1) {\ldots}}
      child {node[ex] (e2) {\ldots}}
    }
    child {
      node[nd] (c3) {\(\mathtt{if}(c_3)\)}
      child {
        node[nd] (c4) {\(\mathtt{if}(c_4)\)}
        child {node[lf] (v1) {\(v_1\)}}
        child {node[lf] (v2) {\(v_2\)}}
      }
      child {node[ex] (e3) {\ldots}}
    };
    %% draw all paths for cont relax
    \edge[cr, red] {c1} {c3};
    \edge[cl, red] {c1} {c2};
    \edge[cr, red] {c2} {e2};
    \edge[cl, red] {c2} {e1};
    \edge[cr, red] {c3} {e3};
    \edge[cl, red] {c3} {c4};
    \edge[cr, red] {c4} {v2};
    \edge[cl, red] {c4} {v1};
    %% single path for reinforce etc.
    \edge[rr, blue, transform canvas={xshift=-0.6mm}] {c1} {c3};
    \edge[rl, blue] {c3} {c4};
    \edge[rr, blue] {c4} {v2};
  \end{tikzpicture}
  \vspace*{-0.2\baselineskip}
  \caption{%
    The challenge faced by continuous-relaxation methods on \glspl{SCFM}---requiring exploration of \textcolor{red}{all branches}, in contrast to exploring only \textcolor{blue}{one branch} at a time.
    Stochastic control flow proceeds through discrete choices (\(c_i\)) yielding values (\(v_i\)).
  }
  \label{fig:exponential-all}
  \vspace*{-2\baselineskip}
\end{figure}
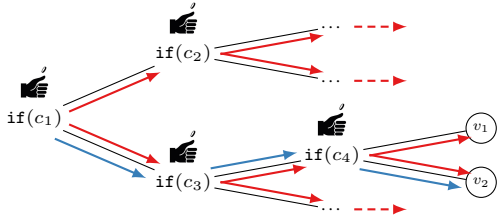

In contrast, \glspl{SCFM} do not lend themselves to continuous-relaxation-based approaches due to the explicit branching requirement on choices from the discrete variables.
Consider for example a simple \gls{SCFM}---a two-mixture \gls{GMM}.
Computing the \gls{ELBO} for this model involves choosing mixture identity (Bernoulli).
Since a sample from a relaxed variable denotes a point on the surface of a probability simplex (e.g. [0.2, 0.8] for a Bernoulli random variable), instead of its vertices (0 or 1), computing the \gls{ELBO} would need evaluation of both branches, weighting the resulting computation under each branch appropriately.
This process can very quickly become intractable for more complex \glspl{SCFM}, as it requires evaluation of \emph{all} possible branches in the computation, of which there may be \emph{exponentially} many, as illustrated in \cref{fig:exponential-all}.

Alternatives to continuous-relaxation methods mainly involve the use of the \gls{IWAE}~\citep{burda2016importance} framework, employing the \acrshort{REINFORCE}~\citep{williams1992simple} gradient estimator, combined with control-variate schemes~\citep{mnih2014neural,mnih2016variational,gu2016muprop,tucker2017rebar,grathwohl2018backpropagation} to help decrease the variance of the na\"ive estimator.
Although this approach ameliorates the problem with continuous relaxations in that it does not require evaluation of all branches, it has other drawbacks.
Firstly, with more particles, the \gls{IWAE} estimator adversely impacts inference-network quality, consequently impeding model learning~\citep{rainforth2018tighter}.
Secondly, its practical efficacy can still be limited due to high variance and the requirement to design and optimize a separate neural network (c.f. \cref{sec:experiments/gmm}).

Having characterized the class of models we are interested in (c.f. \cref{fig:scfm}), and identified a range of current approaches (along with their characteristics) that might apply to such models, we revisit \gls{RWS}~\citep{bornschein2015reweighted}.
Comparing extensively with state-of-the-art methods for learning in \glspl{SCFM}, we
demonstrate its efficacy in learning better generative models and inference networks, using lower variance gradient estimators, over a range of computational budgets.
To this end, we first review state-of-the-art methods for learning deep generative models with discrete latent variables (\cref{sec:background}).
We then revisit \gls{RWS} (\cref{sec:method}) and present an extensive evaluation of these methods (\cref{sec:experiments}) on
\begin{inparaenum}[i)]
\item a \gls{PCFG} model on sentences,
\item the \gls{AIR} model~\citep{eslami2016attend} to perceive and localize multiple \acrshort{MNIST} digits, and
\item a pedagogical \gls{GMM} example that exposes a shortcoming of \gls{RWS} which we then design a fix for. %  using defensive \acrlong{IS}~\citep{hesterberg1995weighted}, and
\end{inparaenum}
Our experiments confirm \gls{RWS} as a competitive, often preferable, alternative for learning \glspl{SCFM}.

\section{BACKGROUND}
\vspace*{-1ex}
\label{sec:background}

Consider data $(x^{(n)})_{n = 1}^N$ sampled from a true (unknown) generative model $p(x)$, a family of generative models $p_\theta(z, x)$ of latent variable $z$ and observation $x$ parameterized by $\theta$ and a family of inference networks $q_\phi(z \given x)$ parameterized by $\phi$.
We aim to learn the generative model by maximizing the marginal likelihood over data: \(\theta^* = \argmax_\theta  \frac{1}{N} \sum_{n = 1}^N  \log p_\theta(x^{(n)})\).
Simultaneously, we would like to learn an inference network $q_\phi(z \given x)$ that amortizes inference given observation $x$; i.e., $q_\phi(z \given x)$ maps an observation $x$ to an approximation of $p_{\theta^*}(z \given x)$.
Amortization ensures this function evaluation is cheaper than performing approximate inference of $p_{\theta^*}(z \given x)$ from scratch.
Our focus here is on such joint learning of generative model and inference network, here referred to as ``learning a deep generative model'', although we note that other approaches exist that learn the generative model~\citep{goodfellow2014generative,mohamed2016learning} or inference network~\citep{paige2016inference,le2017inference} in isolation.

We begin by reviewing \glspl{IWAE}~\citep{burda2016importance} as a general approach for learning deep generative models using \gls{SGD} methods, focusing on generative-model families with discrete latent variables, for which the na\"ive gradient estimator's high variance impedes learning.
We also review control-variate and continuous-relaxation methods for gradient-variance reduction.
\Glspl{IWAE} coupled with such gradient-variance reduction methods are currently the dominant approach for learning deep generative models with discrete latent variables.

\vspace*{-1ex}
\subsection{IMPORTANCE WEIGHTED AUTOENCODERS}
\vspace*{-1ex}

\citet{burda2016importance} introduce the \gls{IWAE}, maximizing the mean \glspl{ELBO} over data, $\frac{1}{N} \sum_{n = 1}^N \ELBO_{\text{IS}}^K(\theta, \phi, x^{(n)})$, where, for $K$ particles,
\begin{align}
  \label{eq:elbo_is}
  \ELBO_{\text{IS}}^K(\theta, \phi, x)
  &= \E_{Q_\phi(z_{1:K} \given x)}
    \!\!\left[ \log\!\left(\!\frac{1}{K} \sum_{k = 1}^K w_k \!\right)\right]\!,\\
  Q_\phi(z_{1:K} \given x)
  &= \prod_{k = 1}^K q_\phi(z_k \given x), \,w_k = \frac{p_\theta(z_k, x)}{q_\phi(z_k \given x)}.
  \nonumber
\end{align}
When $K = 1$, this reduces to the \gls{VAE}~\citep{kingma2014auto,rezende2014stochastic}.
\citet{burda2016importance} show that $\ELBO_{\text{IS}}^K(\theta, \phi, x)$ is a lower bound on $\log p_\theta(x)$ and that increasing~$K$ leads to a tighter lower bound.
Further, tighter lower bounds arising from increasing~$K$ improve learning of the generative model, but impair learning of the inference network~\citep{rainforth2018tighter}, as the signal-to-noise ratio of~\(\theta\)'s gradient estimator is $O(\sqrt{K})$ whereas~\(\phi\)'s is $O(1 / \sqrt{K})$.
Note that although \citet{tucker2019doubly} solve this for reparameterizable distributions, the issue persists for discrete distributions.
Consequently, poor learning of the inference network, beyond a certain point (large~\(K\)), can actually impair learning of the generative model as well; a finding we explore in \cref{sec:experiments/gmm}.

Optimizing the \gls{IWAE} objective using \gls{SGD} methods requires unbiased gradient estimators of $\ELBO_{\text{IS}}^K(\theta, \phi, x)$ with respect to $\theta$ and $\phi$~\citep{robbins1951stochastic}.
$\nabla_\theta \ELBO_{\text{IS}}^K(\theta, \phi, x)$ is estimated by evaluating $\nabla_\theta \log \hat Z_K$ using samples $z_{1:K} \sim Q_\phi(\cdot \given x)$, where $\hat Z_K = \frac{1}{K} \sum_{k = 1}^K\! w_k$.
$\nabla_\phi \!\ELBO_{\acrshort{IS}}^K\!(\theta, \phi, x)$ is estimated similarly for models with reparameterizable latents, discrete (and other non-reparameterizable) latents require the \acrshort{REINFORCE} gradient estimator~\citep{williams1992simple}
\begin{align}
  \!\!\!\!g_{\acrshort{REINFORCE}}
  \!=\! \underbrace{\log \hat Z_K \nabla_\phi \log Q_\phi(z_{1:K} \given x)}_{\circled[5]{1}}
  \!+\! \underbrace{\nabla_\phi \log \hat Z_K}_{\circled[5]{2}}.\!\!
    \label{eq:iwae-reinforce}
\end{align}

\vspace*{-1ex}
\subsection{CONTINUOUS RELAXATIONS AND CONTROL VARIATES}
\vspace*{-1ex}
\label{sec:background/control-variates}

Since the gradient estimator in \cref{eq:iwae-reinforce} typically suffers from high variance, mainly due to the effect of \circled[5]{1}, a number of approaches have been developed to ameliorate the issue.
These can be broadly categorized into approaches that directly transform the discrete latent variables (continuous relaxations), or approaches that target improvement of the na\"ive \acrshort{REINFORCE} estimator (control variates).

\paragraph{Continuous Relaxations:}%
Here, discrete variables are transformed to enable reparameterization~\citep{kingma2014auto,rezende2014stochastic}, helping reduce gradient-estimator variance.
Approaches span the Gumbel distribution~\citep{maddison2017concrete,jang2017categorical}, spike-and-X transforms~\citep{rolfe2016dvae}, overlapping exponentials~\citep{vahdat2018dvaepp}, and generalized overlapping exponentials for tighter bounds~\citep{vahdat2018dvaehash}.

Besides difficulties inherent to such methods, such as tuning temperature parameters, or the suitability of undirected Boltzmann machine priors, these methods are not well suited for learning \glspl{SCFM} as they generate samples on the surface of a probability simplex rather than its vertices.
For example, sampling from a transformed Bernoulli distribution yields samples of the form \([\alpha, (1 - \alpha)]\) rather than simply 0 or 1---the latter form required for branching.
With relaxed samples, as illustrated in \cref{fig:exponential-all}, one would need to execute \emph{all} the exponentially many discrete-variable driven branches in the model, weighting each branch appropriately---something that can quickly become infeasible for even moderately complex models.
However, for purposes of comparison, for relatively simple \glspl{SCFM}, one could apply methods involving continuous relaxations, as demonstrated in \cref{sec:experiments/gmm}.

\paragraph{Control Variates:}%
Here, approaches build on the \acrshort{REINFORCE} estimator for the \gls{IWAE} \gls{ELBO} objective, designing control-variate schemes to reduce the variance of the na\"ive estimator.
\Gls{VIMCO}~\citep{mnih2016variational} eschews designing an explicit control variate, instead exploiting the particle set obtained in \gls{IWAE}.
It replaces \circled[5]{1} with
\begin{align}
\vspace*{-1ex}
  g_{\acrshort{VIMCO}}^{\circled[2]{1}}
  &= \sum_{k = 1}^K (\log \hat Z_K - \Upsilon_{-k}) \nabla_\phi \log q_\phi(z_k \given x),\\[-0.5ex]
  \Upsilon_{-k}
  & = \log \frac{1}{K} \biggl(\exp\biggl(\frac{1}{K - 1} \sum\limits_{\ell \neq k} \log w_\ell\biggr)
    + \sum_{\ell \neq k} w_\ell \biggr) \nonumber
\vspace*{-1ex}
\end{align}
where \(\Upsilon_{-k} \perp\hspace*{-6pt}\perp z_k\) and highly correlated with $\log \hat Z_K$.

Finally, assuming $z_k$ is a discrete random variable with $C$ categories\footnote{The assumption is needed only for notational convenience. However, using more structured latents leads to difficulties in picking the control-variate architecture.}, \acrshort{REBAR}~\citep{tucker2017rebar} and \acrshort{RELAX}~\citep{grathwohl2018backpropagation} improve on \citet{mnih2014neural} and \citet{gu2016muprop}, replacing \circled[5]{1} as
\begin{align}
  g_{\acrshort{RELAX}}^{\circled[2]{1}}
  &= \biggl(\log \hat Z_K - c_\rho(\tilde g_{1:K}) \biggr) \nabla_\phi \log Q_\phi(z_{1:K} \given x) \nonumber \\
  &\quad+ \nabla_\phi c_\rho(g_{1:K}) - \nabla_\phi c_\rho(\tilde g_{1:K}),
\end{align}
where $g_k$ is a $C$-dimensional vector of reparameterized Gumbel random variates, $z_k$ is a one-hot argmax function of $g_k$, and $\tilde g_k$ is a vector of reparameterized conditional Gumbel random variates conditioned on $z_k$.
The conditional Gumbel random variates are a form of Rao-Blackwellization used to reduce variance.
The control variate $c_\rho$, parameterized by $\rho$, is optimized to minimize the gradient variance estimates along with the main \gls{ELBO} optimization, leading to state-of-the-art performance on, for example, sigmoid belief networks~\citep{neal1992connectionist}.
The main difficulty in using this method is choosing a suitable family of $c_\rho$, as some choices lead to higher variance despite concurrent gradient-variance minimization.

\section{REVISITING REWEIGHTED WAKE-SLEEP}
\label{sec:method}

\Acrfull{RWS}~\citep{bornschein2015reweighted} comes from a family of algorithms~\citep{hinton1995wake,dayan1995helmholtz} for learning deep generative models, eschewing a single objective over parameters~\(\theta\) and~\(\phi\) in favour of individual objectives for each.
We review the \gls{RWS} algorithm and discuss its pros and cons.

\subsection{REWEIGHTED WAKE-SLEEP}

% context and overal framework of RWS
\Acrfull{RWS}~\citep{bornschein2015reweighted} is an extension of the \acrlong{WS} algorithm~\citep{hinton1995wake,dayan1995helmholtz} both of which, like \gls{IWAE}, jointly learn a generative model and an inference network given data.
While \gls{IWAE} targets a single objective, \gls{RWS} alternates between objectives, updating the generative model parameters $\theta$ using a \emph{wake-phase $\theta$ update} and the inference network parameters $\phi$ using either a \emph{sleep-} or a \emph{wake-phase $\phi$ update} (or both).

\paragraph{Wake-phase $\theta$ update.}%
Given~$\phi$, $\theta$~is updated using an unbiased estimate of $\nabla_\theta -\big(\frac{1}{N}\!\sum_{n = 1}^N \!\ELBO_{\text{IS}}^K(\theta, \phi, x^{(n)})\!\big)$, obtained without reparameterization or control variates, as the sampling distribution $Q_\phi(\cdot | x)$ is independent of $\theta$\!.%
\footnote{We assume that the deterministic mappings induced by the parameters~\(\theta, \phi\) are themselves differentiable, such that they are amenable to gradient-based learning.}

\paragraph{Sleep-phase $\phi$ update.}%
Here, $\phi$ is updated to minimize the \gls{KL} divergence between the posteriors under the generative model and the inference network, averaged over the data distribution of the current generative model
\begin{align}
  \E_{p_\theta(x)}&[\KL(p_\theta(z \given x), q_\phi(z \given x))] \nonumber\\
  &= \E_{p_\theta(z, x)}[\log p_\theta(z \given x) - \log q_\phi(z \given x)].
  \label{eq:sleep-phi-obj}
\end{align}
Its gradient, $\E_{p_\theta(z, x)}[-\nabla_\phi \log q_\phi(z \given x)]$, is estimated by evaluating $-\nabla_\phi \log q_\phi(z \given x)$, where $z, x \sim p_\theta(z, x)$.
The estimator's variance can be reduced at a standard Monte Carlo rate by increasing the number of samples of $z, x$.

\paragraph{Wake-phase $\phi$ update.}%
\hspace*{-0.5ex}Here, \(\phi\) is updated to minimize the \gls{KL} divergence between the posteriors under the generative model and the inference network, averaged over the true data distribution
\begin{align}
  \hspace*{-0.35em}
  \E_{p(x)}&[\KL(p_\theta(z \given x), q_\phi(z \given x))] \nonumber \\
  &= \E_{p(x)}[\E_{p_\theta(z \given x)}[\log p_\theta(z \given x) - \log q_\phi(z \given x)]]. \label{eq:wake-phi-obj}
\end{align}
The outer expectation $\E_{p(x)}[\E_{p_\theta(z \given x)}[-\nabla_\phi \log q_\phi(z \given x)]]$ of the gradient is estimated using a single sample $x$ from the true data distribution $p(x)$, given which, the inner expectation is estimated using self-normalized importance sampling with $K$ particles, using $q_\phi(z \given x)$ as the proposal distribution.
This results in the following estimator
\begin{align}
  \sum_{k = 1}^K \frac{w_k}{\sum_{\ell = 1}^K w_\ell} \left(-\nabla_\phi \log q_\phi(z_k \given x)\right), \label{eq:wake-phi-est}
\end{align}
where, similar to \cref{eq:elbo_is}, $x \sim p(x)$, $z_k \sim q_\phi(z_k \given x)$, and $w_k = p_\theta(z_k, x) / q_\phi(z_k \given x)$.
Note that \cref{eq:wake-phi-est} is the negative of the second term of the \acrshort{REINFORCE} estimator of the \gls{IWAE} \gls{ELBO} in \cref{eq:iwae-reinforce}.
% \cref{eq:iwae-sampling-dist-weight}.
%
The crucial difference between the \emph{wake-phase $\phi$ update} and the \emph{sleep-phase $\phi$ update} is that the expectation in \cref{eq:wake-phi-obj} is over the \emph{true data distribution} $p(x)$ and the expectation in \cref{eq:sleep-phi-obj} is under the \emph{current model distribution} $p_\theta(x)$.
The former is desirable from the perspective of amortizing inference over data from~\(p(x)\), and although its estimator is biased, this bias decreases as $K$ increases.

\subsection{PROS OF REWEIGHTED WAKE-SLEEP}

While the gradient update of $\theta$ targets the same objective as \gls{IWAE}, the gradient update of $\phi$ targets the objective in \cref{eq:sleep-phi-obj} in the sleep case and \cref{eq:wake-phi-obj} in the wake case.
% This leads to three advantages of \gls{RWS} over \gls{IWAE} for learning inference networks.
This makes \gls{RWS} a preferable option to \gls{IWAE} for learning inference networks because
the $\phi$ updates in \gls{RWS} directly target minimization of the expected \gls{KL} divergences from the true to approximate posterior.
With an increased computational budget, using more Monte Carlo samples in the \emph{sleep-phase $\phi$ update} case and more particles $K$ in the \emph{wake-phase $\phi$ update}, we obtain a better estimator of these expected \gls{KL} divergences.
This is in contrast to \gls{IWAE}, where optimizing $\ELBO_{\acrshort{IS}}^K$ targets a \gls{KL} divergence on an extended sampling space~\citep{le2018autoencoding} which for $K > 1$ doesn't correspond to a \gls{KL} divergence between true and approximate posteriors (in any order).
Consequently, increasing~\(K\) in \gls{IWAE} leads to impaired learning of inference networks~\citep{rainforth2018tighter}.

Moreover, targeting $\KL(p, q)$ as in \gls{RWS} can be preferable to targeting $\KL(q, p)$ as in \glspl{VAE}.
The former encourages \emph{mean-seeking behavior}, having the inference network to put non-zero mass in regions where the posterior has non-zero mass, whereas the latter encourages \emph{mode-seeking behavior}, having the inference network to put mass on one of the modes of the posterior~\citep{minka2005divergence}.
Using the inference network as an \gls{IS} proposal requires mean-seeking behavior~\citep[Theorem 9.2]{owen2013monte}.
Moreover, \citet{chatterjee2018sample} show that the number of particles required for \gls{IS} to accurately approximate expectations of the form $\E_{p(z \given x)}[f(z)]$ is directly related to $\exp(\KL(p, q))$.

\subsection{CONS OF REWEIGHTED WAKE-SLEEP}
\label{sec:disadvantages}

While a common criticism of the wake-sleep family of algorithms is the lack of a unifying objective, we have not found any empirical evidence where this is a problem.
Perhaps a more relevant criticism is that both the sleep and wake-phase $\phi$ gradient estimators are biased with respect to $\nabla_\phi \E_{p(x)}[\KL(p_\theta(z \given x), q_\phi(z \given x))]$.
The bias in the sleep-phase $\phi$ gradient estimator arises from targeting the expectation under the model rather than the true data distribution, and the bias in the wake-phase $\phi$ gradient estimator results from estimating the \gls{KL} divergence using self-normalized \gls{IS}.

In theory, these biases should not affect the fixed point of optimization $(\theta^*\!, \phi^*)$ where $p_{\theta^*}(x) = p(x)$ and $q_{\phi^*}(z \given x) = p_{\theta^*}(z \given x)$.
First, if $\theta \rightarrow \theta^*$ through the wake-phase $\theta$ update, the data distribution bias reduces to zero.
Second, although the wake-phase $\phi$ gradient estimator is biased, it is consistent---with large enough $K$, convergence of stochastic optimization is theoretically guaranteed on convex objectives and empirically on non-convex objectives~\citep{chen2018stochastic}.
Further, this gradient estimator follows the central limit theorem, so its asymptotic variance decreases linearly with $K$~\citep[Eq. (9.8)]{owen2013monte}.
Thus, using larger $K$ improves learning of the inference network.

In practice, the families of generative models, inference networks, and the data distributions determine which of the biases are more significant.
In most of our findings, the bias of the data distribution appears to be the most detrimental.
This is due to the fact that initially $p_\theta(x)$ is quite different from $p(x)$, and hence using sleep-phase $\phi$ updates performs worse than using wake-phase $\phi$ updates.
An exception to this is the \gls{PCFG} experiment (c.f. \cref{sec:experiments/pcfg}) where the data distribution bias is not as large and inference using self-normalized \gls{IS} is extremely difficult.

\vspace*{-1ex}
\section{EXPERIMENTS}
\vspace*{-1ex}
\label{sec:experiments}

The \gls{IWAE} and \gls{RWS} algorithms have primarily been applied to problems with continuous latent variables and/or discrete latent variables that do not actually induce branching (such as sigmoid belief networks;~\cite{neal1992connectionist}).
The purpose of the following experiments is to compare \gls{RWS} to \gls{IWAE} combined with control variates and continuous relaxations (c.f \cref{sec:method}) on models with conditional branching, and show that it outperform such methods.
We empirically demonstrate that increasing the number of particles $K$ can be detrimental in \gls{IWAE} but advantageous in \gls{RWS}, as evidenced by achieved \glspl{ELBO} and average distance between true and amortized posteriors.

In the first experiment, we present learning and amortized inference in a \gls{PCFG}~\citep{booth1973applying}, an example \gls{SCFM} where continuous relaxations are inapplicable.
We demonstrate that \gls{RWS} outperforms \gls{IWAE} with a control variate both in terms of learning and inference.
The second experiment focuses on \glsreset{AIR}\gls{AIR}, the deep generative model of \cite{eslami2016attend}.
It demonstrates that \gls{RWS} leads to better learning of the generative model in a setting with both discrete and continuous latent variables, for modeling a complex visual data domain (c.f. \cref{sec:experiments/air}).
The final experiment involves a \gls{GMM} (\cref{sec:experiments/gmm}), thereby serving as a pedagogical example.
It explains the causes of why \gls{RWS} might be preferable to other methods in more detail.
\footnote{In \cref{app:sigmoid_belief_nets}, we include additional experiments on sigmoid belief networks which, however, are not \glspl{SCFM}.}

Notationally, the different variants of \gls{RWS} will be referred to as \gls{WS} and \gls{WW}.
The \emph{wake-phase $\theta$ update} is always used.
We refer to using it in conjunction with the \emph{sleep-phase $\phi$ update} as \acrshort{WS} and using it in conjunction with the \emph{wake-phase $\phi$ update} as \acrshort{WW}.
Using both \emph{wake-} and \emph{sleep-phase $\phi$ updates} doubles the required stochastic sampling while yielding only minor improvements on the models we considered.
The number of particles $K$ used for the \emph{wake-phase $\theta$} and \emph{$\phi$ updates} is always specified, and computation between them is matched so
a \emph{wake-phase $\phi$ update} with batch size $B$ implies a \emph{sleep phase $\phi$ update} with $KB$ samples.

\subsection{PROBABILISTIC CONTEXT-FREE GRAMMAR}
\label{sec:experiments/pcfg}

\begin{figure*}
  \centering
  \includegraphics[width=\textwidth]{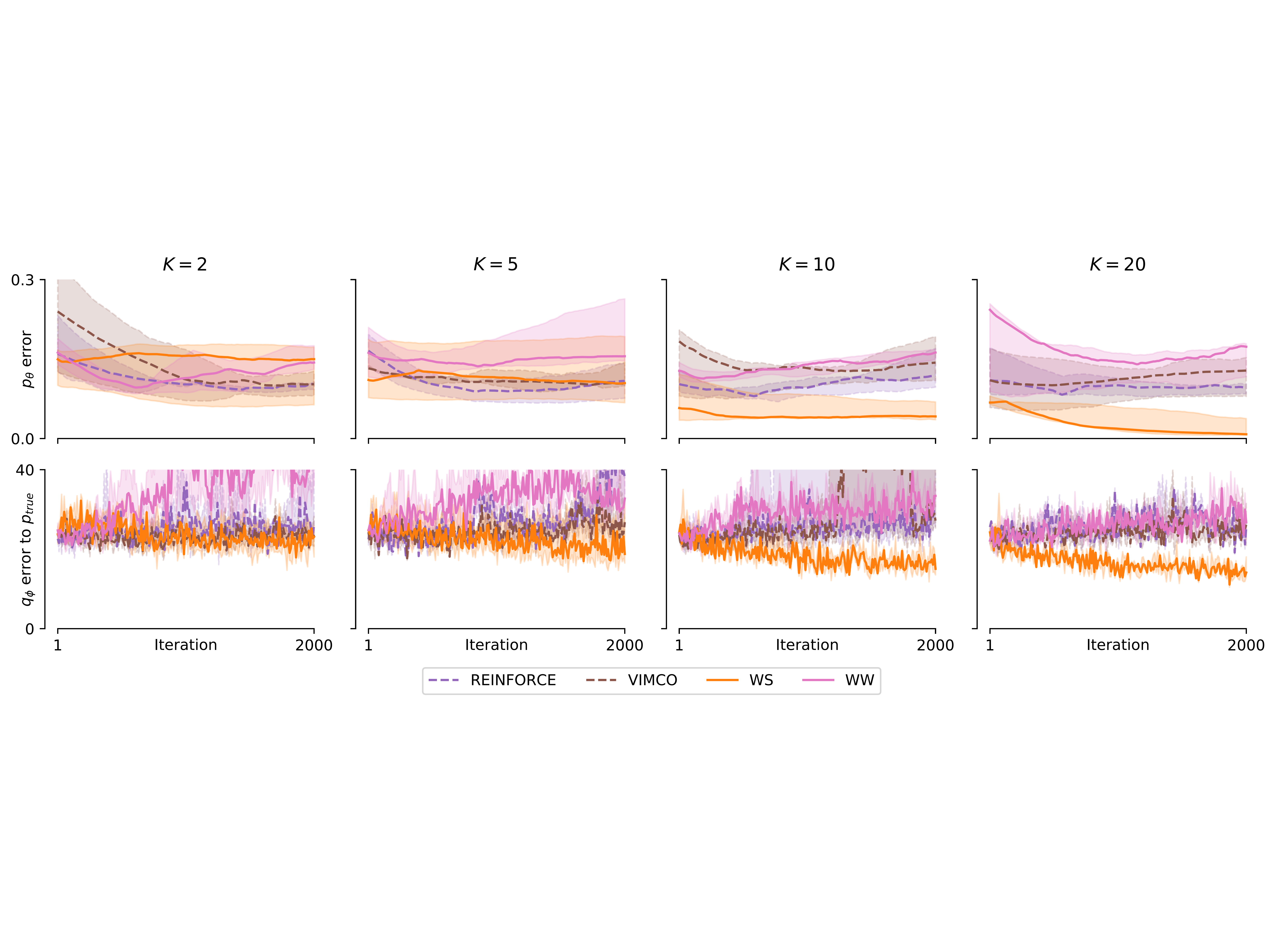}
  \vspace*{-0.7\baselineskip}
  \caption{
    \Gls{PCFG} training.
    \emph{(Top)}
    Quality of the generative model:
    While all methods have the same gradient update for $\theta$, the performance of \gls{WS} improves and is the best as $K$ is increased.
    Other methods, including \gls{WW}, do not yield significantly better model learning as $K$ is increased, since \gls{WS}'s inference network learns the fastest.
    \emph{(Bottom)}
    Quality of the inference network:
    \Gls{VIMCO} and \acrshort{REINFORCE} do not improve with increasing $K$.
    \Gls{WS} performs best as $K$ is increased, and while \gls{WW}'s performance improves, the improvement is not as significant.
    This can be attributed to the data-distribution bias being less significant than the bias coming from self-normalized \gls{IS} (c.f. \cref{sec:disadvantages}).
    Median and interquartile ranges from up to $10$ repeats shown (see text).
  }
  \label{fig:experiments/pcfg/astronomers_errors}
  \vspace*{-2ex}
\end{figure*}

In this experiment we learn model parameters and amortize inference in a \gls{PCFG}~\citep{booth1973applying}.
Each discrete latent variable in a \gls{PCFG} chooses a particular child of a node in a tree.
Depending on each discrete choice, the generative model can lead to different future latent variables.
A \gls{PCFG} is an example of an \gls{SCFM} where continuous relaxations cannot be applied---weighing combinatorially many futures by a continuous relaxation is infeasible and doing so for futures which have infinite latent variables is impossible.

While supervised approaches have recently led to state-of-the-art performance in parsing~\citep{chen2014fast}, \glspl{PCFG} remain one of the key models for unsupervised parsing~\citep{manning1999foundations}.
Learning in a \gls{PCFG} is typically done via expectation-maximization~\citep{dempster1977maximum} which uses the inside-outside algorithm~\citep{lari1990estimation}.
Inference methods are based on dynamic programming~\citep{younger1967recognition,earley1970efficient} or search~\citep{klein2003parsing}.
Applying \gls{RWS} and \gls{IWAE} algorithms to \glspl{PCFG} allows learning from large unlabeled datasets through \gls{SGD} while inference amortization ensures linear-time parsing in the number of words in a sentence, at test-time.
Moreover, using the inference network as a proposal distribution in \gls{IS} provides asymptotically exact posteriors if parses are ambiguous.

% generative model
A \gls{PCFG} is defined by sets of terminals (or words) $\{t_i\}$, non-terminals $\{n_i\}$, production rules $\{n_i \to \zeta_j\}$ with $\zeta_j$ a sequence of terminals and non-terminals, probabilities for each production rule such that $\sum_j~P\left(n_i \!\to\! \zeta_j\right)~=~1$ for each $n_i$, and a start symbol $n_1$.
Consider the \emph{Astronomers \gls{PCFG}} given in \citet[Table
11.2]{manning1999foundations} (c.f. \cref{app:pcfg}).
A parse tree $z$ is obtained by recursively applying the production rules until there are no more non-terminals.
For example, a parse tree (S (NP astronomers) (VP (V saw) (NP stars))) is obtained by applying the production rules as follows:

\par\nobreak\vspace{-1.2em}
{\small
\begin{align*}
  \text{S} \xrightarrow{1.0} \text{NP}\,\text{VP} \xrightarrow{0.1} \text{astronomers}\,\text{VP} \xrightarrow{0.7} \text{astronomers}\,\text{V}\,\text{NP} \\
  \,\xrightarrow{1.0} \text{astronomers}\,\text{saw}\,\text{NP} \xrightarrow{0.18} \text{astronomers}\,\text{saw}\,\text{stars},
\end{align*}
}%
where the probability $p(z)$ is obtained by multiplying the corresponding production probabilities as indicated on top of the arrows.
The likelihood of a \gls{PCFG}, $p(x \given z)$, is $1$ if the sentence $x$ matches the sentence produced by $z$ (in this case ``astronomers saw stars'') and $0$ otherwise.
One can easily construct infinitely long $z$ by choosing productions which contain non-terminals, for example: $\text{S} \to \text{NP}\,\text{VP} \to \text{NP}\,\text{PP}\,\text{VP} \to \text{NP}\,\text{PP}\,\text{PP}\,\text{VP} \to \cdots$.

% p, q architecture
We learn the production probabilities of the \gls{PCFG} and an inference network computing the conditional distribution of a parse tree given a sentence.
The architecture of the inference network is the same as described in \citep[Section 3.3]{le2017inference} except the input to the \gls{RNN} consists only of the sentence embedding, previous sample embedding, and an address embedding.
Each word is represented as a one-hot vector and the sentence embedding is obtained through another \gls{RNN}.
Instead of a hard $\{0, 1\}$ likelihood which can make learning difficult, we use a relaxation, $p(x \given z) = \exp(-L(x, s(z))^2)$, where $L$ is the Levenshtein distance and $s(z)$ is the sentence produced by $z$.
Using the Levenshtein distance can also be interpreted as an instance of approximate Bayesian computation~\citep{sisson2018handbook}.
Training sentences are obtained by sampling from the \emph{astronomers} \gls{PCFG} with the true production probabilities.

% why ws is better?
We run \gls{WW}, \gls{WS}, \gls{VIMCO} and \acrshort{REINFORCE} ten times for $K \in \{2, 5, 10, 20\}$, with batch size \(B=2\), using the Adam optimizer~\citep{kingma2015adam} with default hyperparameters.
We observe that the inference network can often end up sub-optimally sampling very long $z$ (by choosing production rules with many non-terminals), leading to slow and ineffective runs.
We therefore cap the run-time to $100$ hours---out of ten runs, \gls{WW}, \gls{WS}, \gls{VIMCO} and \acrshort{REINFORCE} retain on average $6$, $6$, $5.75$ and $4$ runs respectively
In \cref{fig:experiments/pcfg/astronomers_errors}, we show both
\begin{inparaenum}[(i)]
\item the quality of the generative model as measured by the average \gls{KL} between the true and the model production probabilities, and
\item the quality of the inference network as measured by $\E_{p(x)}[\KL(p(z \given x), q_\phi(z \given x))]$ which is estimated up to an additive constant (the conditional entropy $H(p(z \given x))$) by the sleep-$\phi$ loss \cref{eq:sleep-phi-obj} using samples from the true \gls{PCFG}.
\end{inparaenum}

Quantitatively, \gls{WS} improves as $K$ increases and outperforms \gls{IWAE}-based algorithms both in terms of learning and inference amortization.
While \gls{WW}'s inference amortization improves slightly as $K$ increases, it is significantly worse than \gls{WS}'s.
This is because \gls{IS} proposals will rarely produce a parse tree $z$ for which $s(z)$ matches $x$, leading to extremely biased estimates of the wake-$\phi$ update.
In this case, this bias is more significant than that of the data-distribution which can harm the sleep-$\phi$ update.

\begin{figure}[t]
  \centering
  \includegraphics[width=0.95\linewidth]{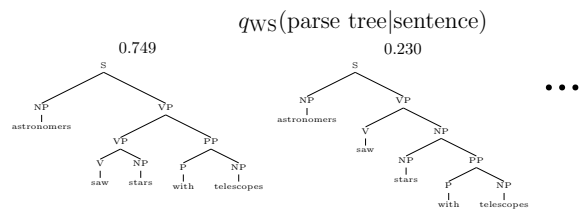}
  \caption{
    Samples from the inference network trained with \gls{WS} ($K = 20$).
    Highest probability samples correspond to correct sentences ($s(z) = x$).
  }
  \label{fig:experiments/pcfg/ws_q_samples}
  \vspace*{-2ex}
\end{figure}

We inspect the quality of the inference network by sampling from it.
\Cref{fig:experiments/pcfg/ws_q_samples}, shows samples from an inference network trained with \gls{WS}, conditioned on the sentence ``astronomers saw stars with telescopes'', weighted according to the frequency of occurrence.
\Cref{app:pcfg} further includes samples from an inference network trained with \gls{VIMCO}, showing that none of them match the given sentence ($s(z) \neq x$), and whose production probabilities are poor, unlike with \gls{RWS}.

\begin{figure*}[!ht]
  \includegraphics[width=0.33\textwidth]{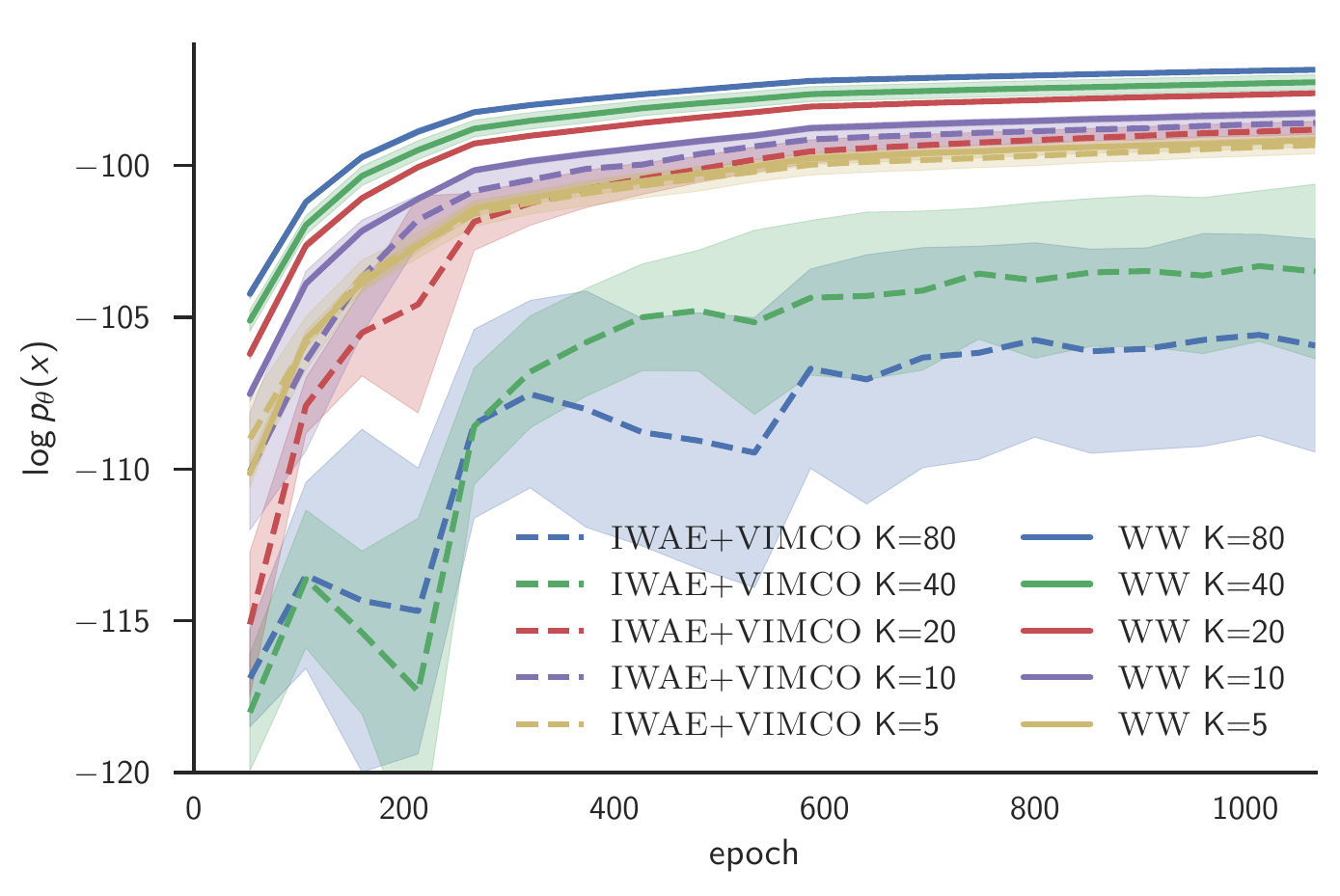}
  \includegraphics[width=0.33\textwidth]{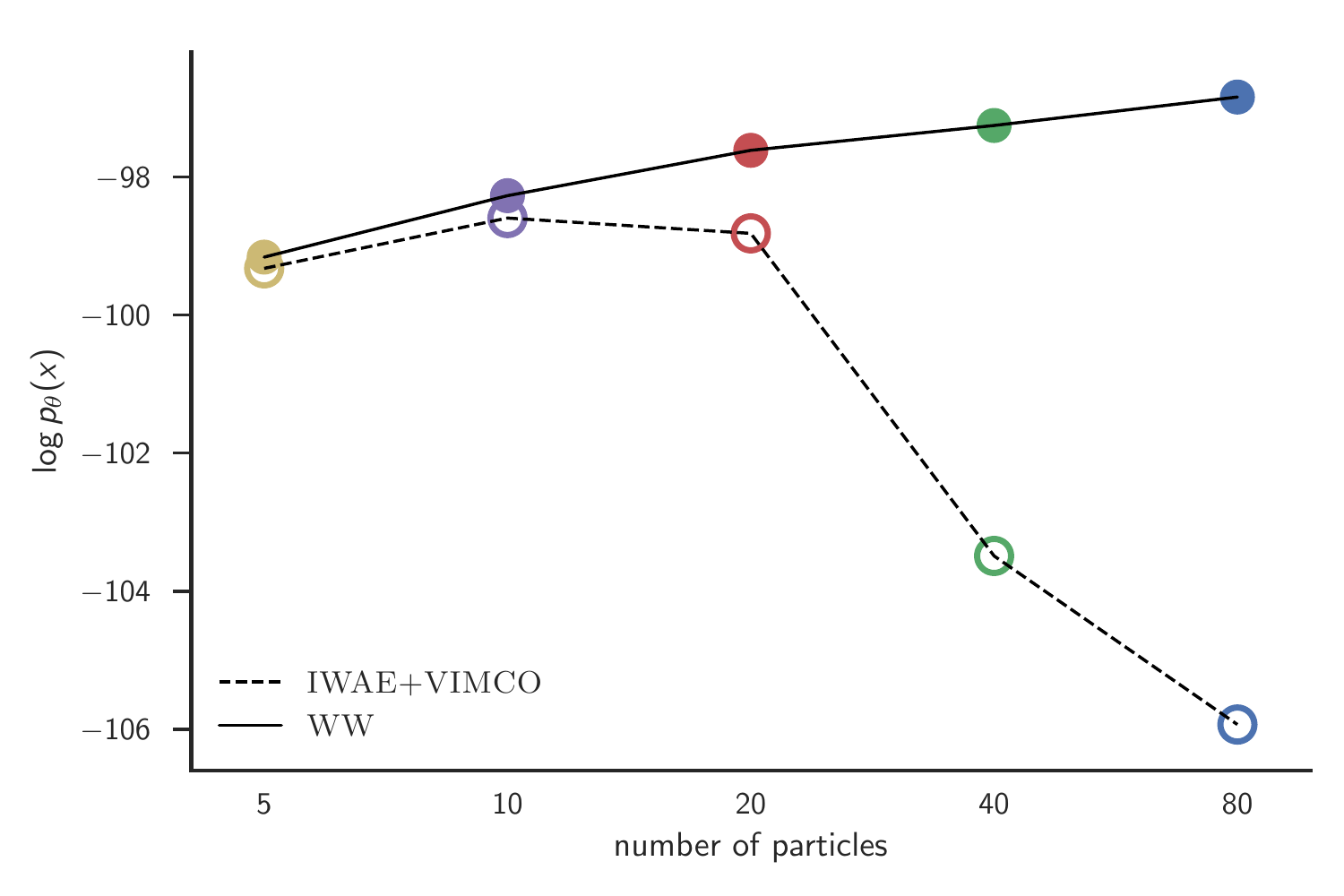}
  \includegraphics[width=0.33\textwidth]{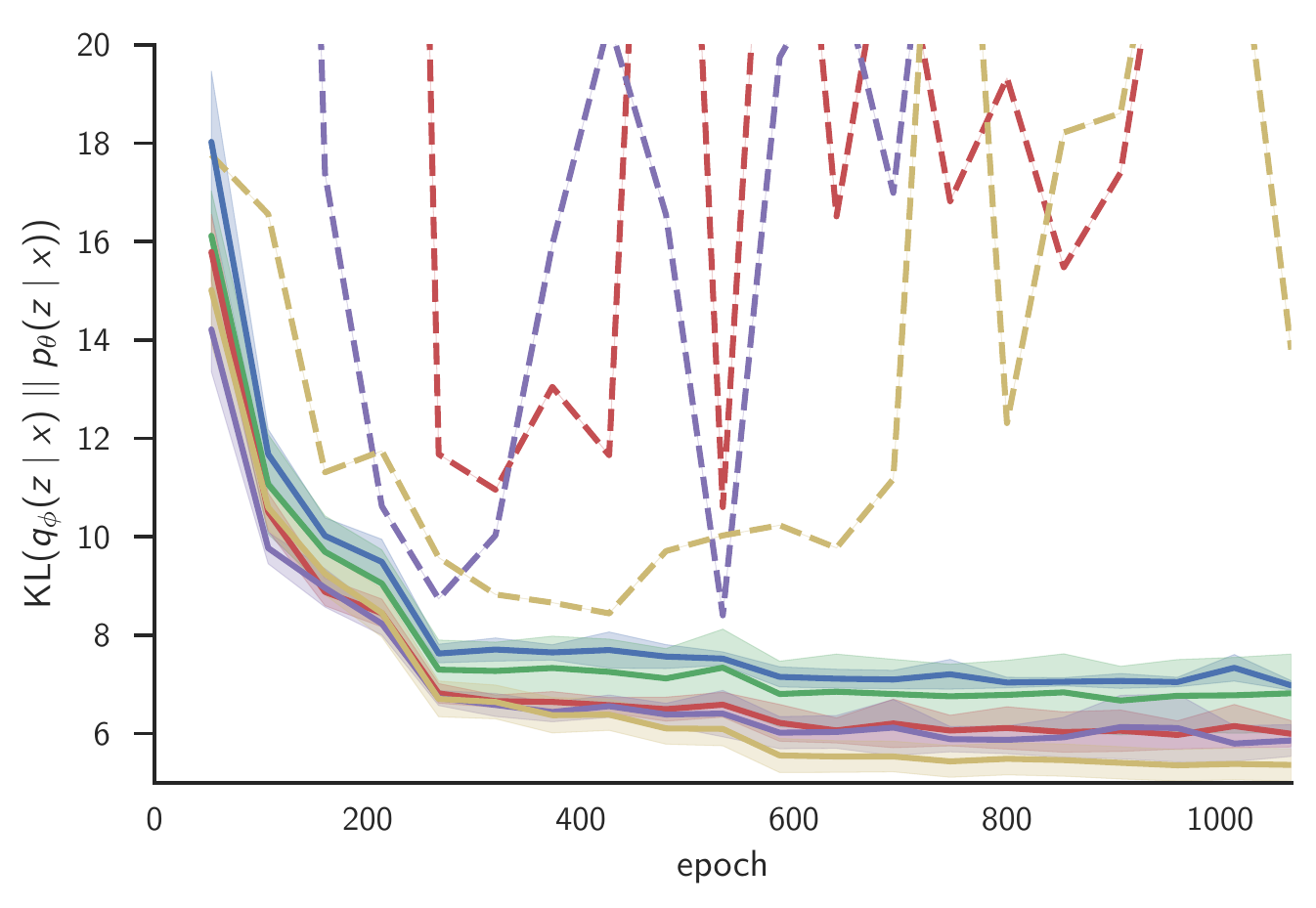}
  \caption{
    Training of \gls{AIR}.
    \emph{(Left)}
    Training curves:
    training with \gls{VIMCO} leads to larger variance in training than \gls{WW}.
    \emph{(Middle)}
    Log evidence values at the end of training:
    increasing number of particles improves \gls{WW} monotonically but improves \gls{VIMCO} only up to a point ($K = 10$ is the best).
    \emph{(Right)}
    \gls{WW} results in significantly lower variance and better inference networks than \gls{VIMCO}.
    Note that \gls{KL} is between the inference network and the \emph{current} generative model.
  }
  \label{fig:air_logp}
  \vspace*{-2ex}
\end{figure*}

\subsection{ATTEND, INFER, REPEAT}
\label{sec:experiments/air}

Next, we evaluate \gls{WW} and \gls{VIMCO}
on \gls{AIR}~\citep{eslami2016attend}, a structured deep generative model with both discrete and continuous latent variables.
\Gls{AIR} uses the discrete variable to decide how many continuous variables are necessary to explain an image.
The sequential inference procedure of \gls{AIR} poses a difficult problem, since it implies a sequential decision process with possible branching.
See \citep{eslami2016attend} and \cref{app:air} for the model notation and details.

% training
We set the maximum number of inference steps in \gls{AIR} to three and train on $50 \times 50$ images with zero, one or two \acrshort{MNIST} digits.
The training and testing data sets consist of $60000$ and $10000$ images respectively, generated from the respective \acrshort{MNIST} train/test datasets.
Unlike \gls{AIR}, which used Gaussian likelihood with fixed standard deviation and continuous inputs (i.e., input $\mathbf{x} \in [0, 1]^{50 \times 50}$), we use a Bernoulli likelihood and binarized data; the stochastic binarization is similar to \citet{burda2016importance}.
Training is performed over two million iterations by RmsProp \citep{tieleman2012rms} with the learning rate of $10^{-5}$, which is divided by three after $400$k and $1000$k training iterations.
We set the glimpse size to $20 \times 20$.

% evaluation
We first evaluate the generative model via the average test log marginal where each log marginal is estimated by a one-sample, $5000$-particle \gls{IWAE} estimate.
The inference network is then evaluated via the average test \gls{KL} from the inference network to the posterior under the current model where each $\KL(q_\phi(z \given x), p_\theta(z \given x))$ is estimated as a difference between the log marginal estimate above and a $5000$-sample, one-particle \gls{IWAE} estimate.
Note that this estimate is just a proxy to the desired \gls{KL} from the inference network to the \emph{true} model posterior.

% result
This experiment confirms that increasing number of particles improves \gls{VIMCO} only up to a point, whereas \gls{WW} improves monotonically with increased~\(K\) (\cref{fig:air_logp}).
\Gls{WW} also results in significantly lower variance and better inference networks than \gls{VIMCO}.

\subsection{GAUSSIAN MIXTURE MODEL}
\label{sec:experiments/gmm}

\begin{figure*}[!ht]
  \centering
  \includegraphics[width=\textwidth]{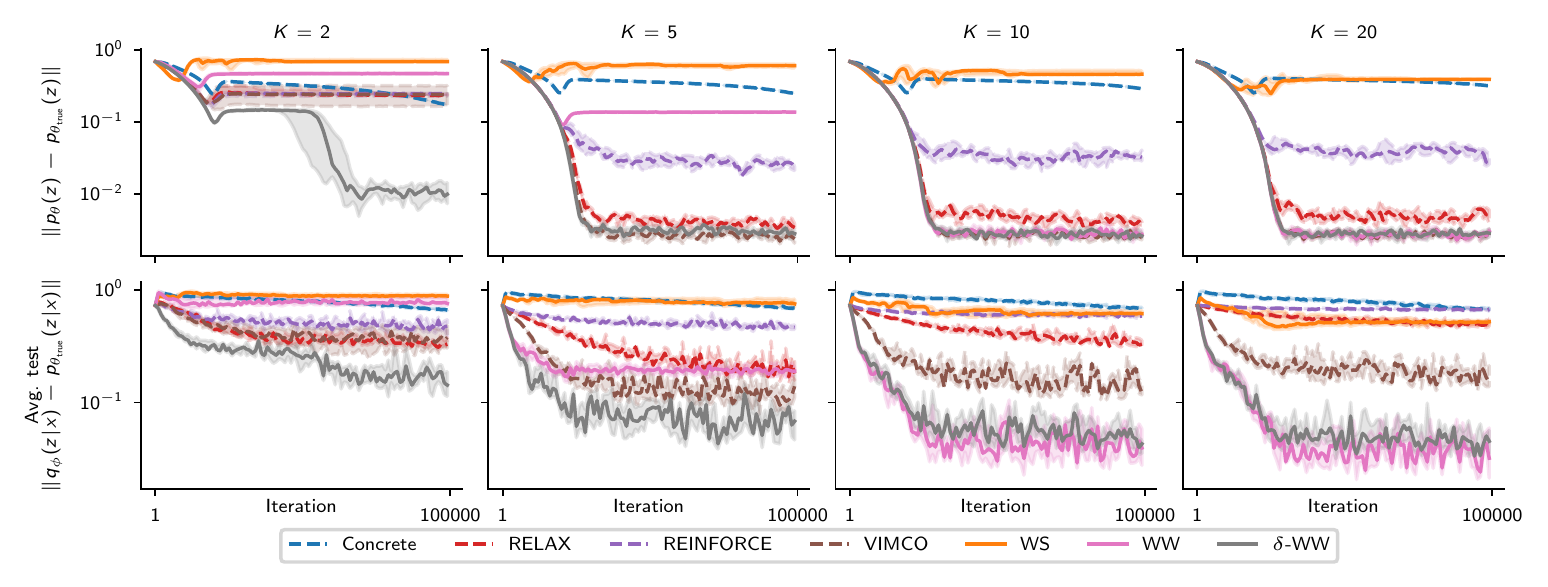}
  \vspace*{-4ex}
  \caption{
    \Gls{GMM} training.
    Median and interquartile ranges from $10$ repeats shown.
    \emph{(Top)}
    Quality of the generative model:
    \gls{WS} and \gls{WW} improve with more particles thanks to lower variance and lower bias estimators of the gradient respectively.
    \Gls{IWAE} methods suffer with a larger particle budget~\citep{rainforth2018tighter}.
    \Gls{WS} performs the worst as a consequence of computing the expected \gls{KL} under the model distribution~\(p_\theta(x)\)~\cref{eq:sleep-phi-obj} instead of the true data distribution~\(p(x)\) as with \gls{WW}~\cref{eq:wake-phi-obj}.
    \Gls{WW} suffers from branch-pruning (see text) in low-particle regimes, but learns the best model fastest in the many-particle regime; $\delta$-\gls{WW} additionally learns well in the low-particle regime.
    \emph{(Bottom)}
    Both inference network and generative model quality develop identically.
  }
  \label{fig:gmm}
  \vspace*{-1ex}
\end{figure*}

In order to examine the differences between \gls{RWS} and \gls{IWAE} more closely, we study  a \gls{GMM} which branches on a discrete latent variable to select cluster assignments.
The generative model and inference network are defined as
%
% {\footnotesize
\begin{align*}
  p_\theta(z) &= \mathrm{Cat}(z \given \mathrm{softmax}(\theta)),\,\,
  p(x \given z) = \mathcal{N}(x \given \mu_z, \sigma_z^2), \\
  q_\phi(z \given x) &= \mathrm{Cat}(z \given \mathrm{softmax}(\eta_\phi(x))),
\end{align*}
% }
%
where $z \in \{0, \dotsc, C - 1\}$,  $C$ is the number of clusters and $\mu_c, \sigma_c^2$ are fixed to $\mu_c = 10c$ and $\sigma_c^2 = 5^2$.
The generative model parameters are $\theta \in \mathbb R^C$.
The inference network consists of a \acrlong{MLP} $\eta_\phi: \mathbb R \to \mathbb R^C$, with the $1$-$16$-$C$ architecture and the $\tanh$ nonlinearity, parameterized by $\phi$.
The chosen family of inference networks is empirically expressive enough to capture the posterior under the true model.
The true model is set to $p_{\theta_\text{true}}(x)$ where $\mathrm{softmax}(\theta_\text{true})_c = (c + 5) / \sum_{i = 1}^C (i + 5)$ ($c = 0, \dotsc, C - 1$), i.e. the mixture probabilities are linearly increasing with the $z$
.
% (\cref{fig:true-gmm}).
We fix the mixture parameters in order to study the important features of the problem at hand in isolation.

% training details
We train using \gls{WS}, \gls{WW}, as well as using \gls{IWAE} with \acrshort{REINFORCE}, \acrshort{RELAX}, \gls{VIMCO} and the Concrete distribution.
We attempted different variants of relaxations \citep{rolfe2016dvae,vahdat2018dvaepp} in this setting, but they performed considerably worse than any of the alternatives (c.f. \Cref{app:dvae}).
We fix $C = 20$ and increase number of particles from $K = 2$ to $20$.
We use the Adam optimizer with default parameters.
Each training iteration samples a batch of $100$ data points from the true model.
Having searched over several temperature schedules for the Concrete distribution, we use the one with the lowest trainable terminal temperature (linearly annealing from $3$ to $0.5$).
We found that using the control variate $c_\rho(g_{1:K}) = \frac{1}{K} \sum_{k = 1}^K \acrshort{MLP}_\rho([x, g_k])$, with \gls{MLP} architecture $(1 + C)$-$16$-$16$-$1$ ($\tanh$) led to most stable training (c.f.\ \cref{app:gmm}).

% evaluation details
The generative model is evaluated via the $L_2$ distance between the \glspl{PMF} of its prior and true prior as $\|\mathrm{softmax}(\theta) - \mathrm{softmax}(\theta_\text{true})\|$.
The inference network is evaluated via the $L_2$ distance between \glspl{PMF} of the current and true posteriors, averaged over a fixed set~(\(M=100\)) of observations $(x_{\text{test}}^{(m)})_{m = 1}^{M}$ from the true model: $\frac{1}{M} \sum_{m = 1}^M \|q_\phi(z \given x_{\text{test}}^{(m)}) - p_{\theta_{\text{true}}}(z \given x_{\text{test}}^{(m)})\|$.

We demonstrate that using \gls{WS} and \gls{WW} with larger particle budgets leads to better inference networks whereas this is not the case for \gls{IWAE} methods (\cref{fig:gmm}, bottom).
Recall that the former is because using more samples to estimate the gradient of the sleep $\phi$ objective \cref{eq:sleep-phi-obj} for \gls{WS} reduces variance at a standard Monte Carlo rate and that using more particles in \cref{eq:wake-phi-est} to estimate the gradient of the wake $\phi$ objective results in a lower bias.
The latter is because using more particles results in the signal-to-noise of \gls{IWAE}'s $\phi$ gradient estimator to drop at the rate $O(1 / \sqrt{K})$~\citep{rainforth2018tighter}.

Learning of the generative model, through inference-network learning, also monotonically improves with increasing \(K\) for \gls{WS} and \gls{WW}, but worsens for all \gls{IWAE} methods except \gls{VIMCO}, since the $\theta$ gradient estimator (common to all methods), $\nabla_\theta \ELBO_{\acrshort{IS}}^K(\theta, \phi, x)$ can be seen as an importance sampling estimator whose quality is tied to the proposal distribution (inference network).

To highlight the difference between \gls{WW} and \gls{WS}, we study the performance of the generative model and the inference network for different initializations of $\theta$.
In \cref{fig:gmm}, $\theta$ is initialized such that the mixture probabilities are exponentially decreasing with $z$ which results in the data distribution $p_\theta(x)$ being far from $p_{\theta^*}(x)$.
Consequently, the sleep-phase $\phi$ update is highly biased which is supported by \gls{WS} being worse than \gls{WW}.
On the other hand, if $\theta$ is initialized such that the mixture probabilities are equal, $p_\theta(x)$ is closer to $p_{\theta^*}(x)$, which is supported by \gls{WS} outperforming \gls{WW} (see \cref{app:gmm/additional}).

\begin{figure*}[!ht]
  \centering
  \includegraphics[width=\textwidth]{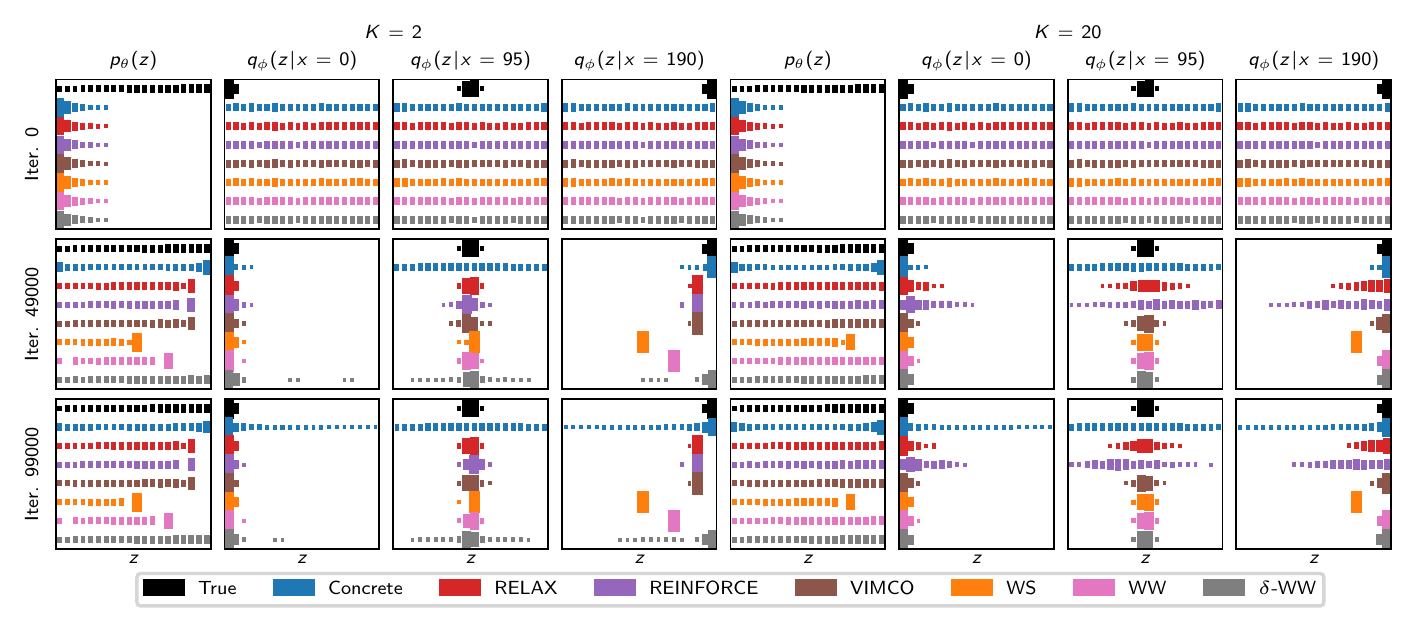}
  \vspace*{-4ex}
  \caption{
    Generative model and inference network during \gls{GMM} training shown as Hinton diagrams where areas are proportional to probability.
    Rows correspond to start, middle and end of optimization.
    \emph{(Left half)}
    Learning with few particles leads to the branch-pruning (described in text) of the inference network (shown as conditional \gls{PMF} given different $x$) and the generative model (first column of each half) for all methods except $\delta$-\gls{WW}.
    Concrete distribution fails.
    \emph{(Right half)}
    Learning with many particles leads to branch-pruning only for \gls{WS}; \gls{WW} and $\delta$-\gls{WW} succeed where \gls{IWAE} fails, learning a suboptimal final generative model.
  }
  \label{fig:gmm2}
\end{figure*}

We now describe a failure mode affecting \gls{WS}, \gls{WW}, \gls{VIMCO}, \acrshort{RELAX} and \acrshort{REINFORCE} due the adverse initialization of $\theta$ which we call \emph{branch-pruning}.
It is best illustrated by inspecting the generative model and the inference network as training progresses, focusing on the low-particle ($K = 2$) regime (\cref{fig:gmm2}).
For \gls{WS}, the generative model $p_\theta(z)$ peaks at $z = 9$ and puts zero mass for $z > 9$; the inference network $q_\phi(z \given x)$ becomes the posterior for this model which, here, has support at most $\{0, \dotsc, 9\}$ for all $x$.
This is a local optimum for \gls{WS} as
\begin{inparaenum}[(i)]
\item the inference network already approximates the posterior of the model $p_\theta(z, x)$ well, and
\item the generative model $p_\theta(z)$, trained using samples from $q_\phi(z \given x)$, has no samples outside of its current support.
\end{inparaenum}
Similar failures occur for \gls{WW} and \gls{VIMCO}/\acrshort{RELAX}/\acrshort{REINFORCE} although the support of the locally optimal $p_\theta(z)$ is larger ($\{0, \dotsc, 14\}$ and $\{0, \dotsc, 17\}$ respectively).

While this failure mode is a particular feature of the adverse initialization of $\theta$, we hypothesize that \gls{WS} and \gls{WW} suffer from it more, as they alternate between two different objectives for optimizing $\theta$ and $\phi$.
\Gls{WS} attempts to amortize inference for the current model distribution $p_\theta(x)$ which reinforces the coupling between the generative model and the inference network, making it easier to get stuck in a local optimum.
\Gls{WW} with few particles (say $K = 1$) on the other hand, results in a highly-biased gradient estimator \cref{eq:wake-phi-est} that samples $z$ from $q_\phi(\cdot \given x)$ and evaluates $\nabla_\phi \log q_\phi(z \given x)$; this encourages the inference network to concentrate mass.
This behavior is not seen in \gls{WW} with many particles where it is the best algorithm at learning both a good generative model and inference network (\cref{fig:gmm}; \cref{fig:gmm2}, right).

We propose a simple extension of \gls{WW}, denoted $\delta$-\acrshort{WW}, that mitigates this shortcoming by changing the proposal of the self-normalized importance sampling estimator in \cref{eq:wake-phi-est} to \(q_{\phi, \delta}(z \given x) = (1 - \delta) q_\phi(z \given x) + \delta \mathrm{Uniform}(z)\).
We use $\delta = 0.2$, noting that the method is robust to a range of values.
Using a different proposal than the inference network $q_\phi(z \given x)$ means that using the low-particle estimator in \cref{eq:wake-phi-est} no longer leads to branch-pruning.
This is known as defensive \acrlong{IS}~\citep{hesterberg1995weighted}, and is used to better estimate integrands that have long tails using short-tailed proposals.
Using $\delta$-\gls{WW} outperforms all other algorithms in learning both the generative model and the inference network in the low-$K$ regime and performs similarly as \gls{WW} in the high-$K$ regime.

\section{DISCUSSION}
\label{sec:discussion}

The central argument here is that where one needs both amortization and model learning for \glspl{SCFM}, the \gls{RWS} family of methods is preferable to \gls{IWAE} with either continuous relaxations or control-variates.
The \gls{PCFG} experiment (\cref{sec:experiments/pcfg}) demonstrates a setting where continuous relaxations are inapplicable due to potentially infinite recursion, but where \gls{RWS} applies and \gls{WS} outperforms all other methods.
The \gls{AIR} experiment (\cref{sec:experiments/air}) highlights a case where with more particles, performance of \gls{VIMCO} degrades for the inference network~\citep{rainforth2018tighter} and consequently the generative model as well, but where \gls{RWS}'s performance on both increases monotonically.
Finally, the analysis on \glspl{GMM} (\cref{sec:experiments/gmm}) focuses on a simple model to understand nuances in the performances of different methods.
Beyond implications from prior experiments, it indicates that for the few-particle regime, the \gls{WW} gradient estimator can be biased, leading to poor learning.
For this, we design an alternative involving defensive sampling that ameliorates the issue.
The precise choice of which variant of \gls{RWS} to employ depends on which of the two kinds of gradient bias described in \cref{sec:disadvantages} dominates.
Where the data distribution bias dominates, as with the \gls{AIR} experiment, \gls{WW} is preferable, and where the self-normalized \gls{IS} bias dominates, as in the \gls{PCFG} experiment, \gls{WS} is preferable.
In the \gls{GMM} experiment, we verify this empirically by studying two optimization procedures with low and high data distribution biases.

\subsubsection*{Acknowledgments}

TAL's research leading to these results is supported by EPSRC DTA and Google (project code DF6700) studentships.
AK's and YWT's research leading to these results are supported by funding from the European Research Council under the European Union’s Seventh Framework Programme (FP7/2007-2013) ERC grant agreement no. 617071.
NS is supported by EPSRC/MURI grant EP/N019474/1.
FW's research leading is supported by The Alan Turing Institute under the EPSRC grant EP/N510129/1; DARPA PPAML through the U.S. AFRL under Cooperative Agreement FA8750-14-2-0006; Intel and DARPA D3M, under Cooperative Agreement FA8750-17-2-0093.

\clearpage
\newpage

\renewcommand*{\bibfont}{\small}
\bibliography{main}
\bibliographystyle{plainnat}
\balance

\clearpage
\newpage
\appendix

\section{PROBABILISTIC CONTEXT-FREE GRAMMAR}
\label{app:pcfg}

We show the \emph{astronomers} \gls{PCFG} in \cref{fig:app/pcfg/astronomers}.
\Cref{fig:experiments/pcfg/vimco_q_samples} shows samples from an inference network trained with \gls{VIMCO} with $K = 20$, conditioned on the sentence $x = $ ``astronomers saw stars with telescopes''.
\Cref{fig:experiments/pcfg/production_probs} shows production probabilities of the non-terminal NP learned by \gls{VIMCO} and \gls{WS} with $K = 20$.
\begin{figure}[htb]
  \begin{align*}
    \text{S} \to&\,\, \text{NP}\,\text{VP}\,(1.0) \\
    \text{NP} \to&\,\, \text{NP}\,\text{PP}\,(0.4) | \text{astronomers}\,(0.1) | \text{ears}\,(0.18) | \\
    &\,\, \text{saw}\,(0.04) | \text{stars}\,(0.18) | \text{telescopes}\,(0.1) \\
    \text{VP} \to&\,\, \text{V}\,\text{NP}\,(0.7) | \text{VP}\,\text{PP}\,(0.3) \\
    \text{PP} \to&\,\, \text{P}\,\text{NP}\,(1.0) \\
    \text{P} \to&\,\, \text{with}\,(1.0) \\
    \text{V} \to&\,\, \text{saw}\,(1.0).
  \end{align*}
  \caption{\emph{The astronomers \acrshort{PCFG}} from \citet[Table 11.2]{manning1999foundations}. The terminals are $\{\text{astronomers}, \text{ears}, \text{saw}, \text{stars}, \text{telescopes}, \text{with}\}$, the non-terminals are $\{\text{S}, \text{NP}, \text{VP}, \text{PP}, \text{P}, \text{V}\}$ and the start symbol is S.
  Each row above lists production rules $\{n_i \to \zeta_j\}$ with the corresponding probabilities $p_{ij}$ in the format $n_i \to \zeta_1\,(p_{i1}) | \zeta_2\,(p_{i2}) | \cdots | \zeta_J\,(p_{iJ})$.}
  \label{fig:app/pcfg/astronomers}
\end{figure}
\begin{figure}[htb]
  \includegraphics[scale=0.6]{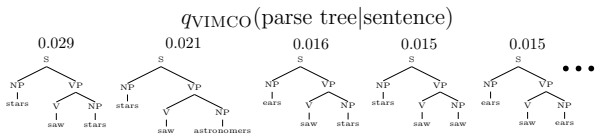}
  \caption{Samples from the inference network which was trained with \gls{VIMCO} with $K = 20$.}
  \label{fig:experiments/pcfg/vimco_q_samples}
  \vspace*{-2ex}
\end{figure}
\begin{figure}[htb]
  \centering
  \includegraphics[width=0.7\linewidth]{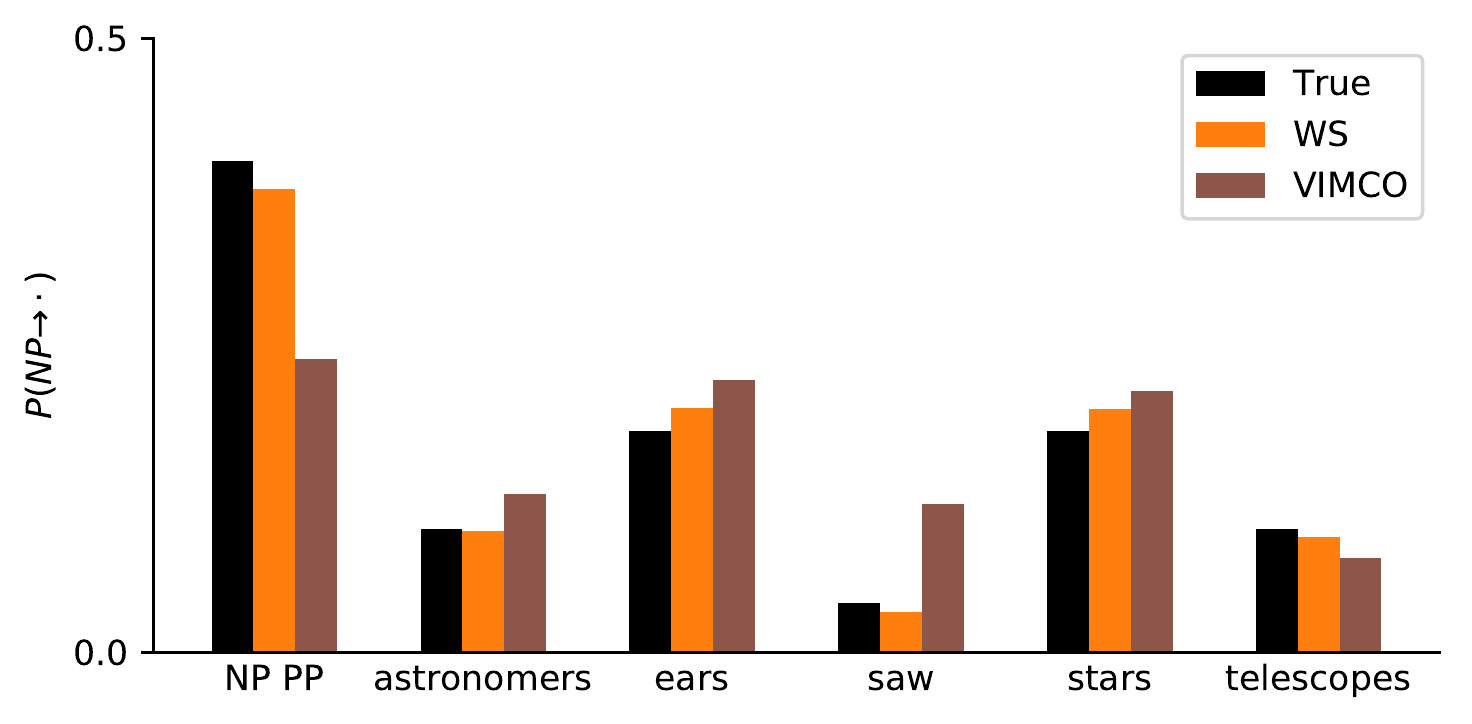}
  \caption{Production probabilities for the non-terminal NP learned via \gls{WS} and \gls{VIMCO} with $K = 20$.}
  \label{fig:experiments/pcfg/production_probs}
  \vspace*{-2ex}
\end{figure}

\section{ATTEND, INFER, REPEAT}
\label{app:air}

\Gls{AIR} is a model with many components and might be difficult to understand if not described explicitly.
Here, we outline details of our implementation and provide pseudo-code for the inference (\cref{algo:air_inference}) and generative models (\cref{algo:air_generation}) in the case of continuous data and Gaussian data likelihood.

\begin{algorithm}[!h]
    \caption{Inference in \gls{AIR}}
    \label{algo:air_inference}
    \DontPrintSemicolon
    % \SetAlgoLined
    \SetKwInOut{Input}{Input}
    \SetKwInOut{Output}{Output}
    \SetSideCommentLeft
    \Input{Image $\bm{x}$,\\ maximum number of inference steps $N$}
    $\bm{h}_0, \bm{z}^\text{what}_0, \bm{z}^\text{where}_0$ = initialize()\\
    \For{$n \in [1, \dots, N]$}{
        $\bm{w}_n, \bm{h}_n = R_\phi \left( \bm{x}, \bm{z}^\text{what}_{n-1}, \bm{z}^\text{where}_{n-1}, \bm{h}_{n-1} \right)$\\
        $p_n \sim \mathrm{Bernoulli} (p \mid \bm{w}_n)$\\
        \If{$p_n = 0$}{
            break
        }
        $\bm{z}^\text{where}_n \sim q_\phi^\text{where} \left( \bm{z}^\text{where} \mid \bm{w}_n \right)$\\
        $\bm{g}_n = \text{STN} \left( \bm{x}, \bm{z}^\text{where}_n \right)$\\
        $\bm{z}^\text{what}_n \sim q_\phi^\text{what} \left( \bm{z}^\text{what} \mid \bm{g}_n \right)$\\
    }
\Output{$\bm{z}^\text{what}_{1:n}$, $\bm{z}^\text{where}_{1:n}$, $n$}
\end{algorithm}
\begin{algorithm}[!h]
    \caption{Generation in \gls{AIR}}
    \label{algo:air_generation}
    \DontPrintSemicolon
    \SetKwInOut{Input}{Input}
    \SetKwInOut{Output}{Output}
    \SetSideCommentLeft
    \Input{$\bm{z}^\text{what}_{1:n}$, $\bm{z}^\text{where}_{1:n}$, $n$}
    $\bm{y}_0 = \bm{0}$\\
    \For{$t \in [1, \dots, n]$}{
        $\hat{\bm{g}}_t = h_\theta^\text{dec} \left( \bm{z}^\text{what}_t \right)$\\
        $\bm{y}_t = \bm{y}_{t-1} + \text{STN}^{-1} \left( \hat{\bm{g}}_t, \bm{z}^\text{where}_t \right)$\\
    }
    $\hat{\bm{x}} \sim \mathrm{Normal} \left(\bm{x} \mid \bm{y}_n, \sigma^2_x \bm{I} \right)$\\
    \Output{$\hat{\bm{x}}$}
\end{algorithm}
\vspace*{-2ex}

\section{GAUSSIAN MIXTURE MODEL}
\label{app:gmm}

\subsection{CONTROL VARIATES}
Here, we present the architectures for the \acrshort{REBAR}/\acrshort{RELAX} control variate used in the \gls{GMM} experiment.

The reparameterized sampling of Gumbels and conditional Gumbels is described by \citet[Appendix C]{tucker2017rebar} and \citet[Appendix B]{grathwohl2018backpropagation}.
In the following, we describe architectures used for the \gls{GMM} experiment (\cref{sec:experiments/gmm}).

\Acrshort{REBAR} proposes the following architecture for the control variate $c_\rho(g_{1:K})$:
\begin{align}
    c_\rho^{\acrshort{REBAR}}(g_{1:K}) = \rho_1 \log\left(\frac{1}{K} \sum_{k = 1}^K \frac{p_\theta(\mathrm{sm}(g_k / e^{\rho_2}), x)}{q_\phi(\mathrm{sm}(g_k / e^{\rho_2}) \given x)}\right), \label{eq:rebar-c}
\end{align}
where $\rho = (\rho_1, \rho_2)$, and $\mathrm{sm}$ refers to the softmax function.
While the functional form of \cref{eq:rebar-c} suggests that it will be highly correlated with $\log(\frac{1}{K} \sum_{k = 1}^K w_k)$, the terms \glspl{PMF} in the fraction are undefined due to the softmax.
A straightforward fix is to evaluate ``a soft \gls{PMF}'' instead:
\begin{align*}
    &p_\theta(\mathrm{sm}(g_k / e^{\rho_2}), x) \\
    &= p_\theta(\mathrm{sm}(g_k / e^{\rho_2})) p(x \given \mathrm{sm}(g_k / e^{\rho_2})) \\
    &= \mathrm{Categorical}(\mathrm{sm}(g_k / e^{\rho_2}) \given \mathrm{sm}(\theta)) \cdot \\
    & \qquad\mathrm{Normal}(x \given \mu_{\mathrm{sm}(g_k / e^{\rho_2})}, \sigma_{\mathrm{sm}(g_k / e^{\rho_2})}^2) \\
    &\approx  \mathrm{sm}(g_k / e^{\rho_2})^\intercal \mathrm{sm}(\theta) \cdot \\
    & \qquad \mathrm{Normal}(x \given \mu^\intercal \mathrm{sm}(g_k / e^{\rho_2}), (\sigma^2)^\intercal \mathrm{sm}(g_k / e^{\rho_2})), \\
    &q_\phi(\mathrm{sm}(g_k / e^{\rho_2}) \given x) \\
    &= \mathrm{Categorical}(\mathrm{sm}(g_k / e^{\rho_2}) \given \mathrm{sm}(\eta_\phi(x))) \\
    &\approx \mathrm{sm}(g_k / e^{\rho_2})^\intercal \mathrm{sm}(\eta_\phi(x)).
\end{align*}
Optimization of the log-temperature $\rho_2$ is highly sensitive as low values can make the training unstable.

\Acrshort{RELAX} proposes using an arbitrary neural network for $c_\rho(g_{1:K})$.
Due to the symmetry in the arguments, we pick the following for the \gls{GMM} experiment:
\begin{align}
    c_\rho^{\acrshort{RELAX}}(g_{1:K}) = \frac{1}{K} \sum_{k = 1}^K \acrshort{MLP}_\rho([x, g_k]), \label{eq:relax-c}
\end{align}
where the architecture of the \gls{MLP} is $(1 + C)$-$16$-$16$-$1$ (with the $\tanh$ nonlinearity between layers) and $\rho$ are the weights are its weights.
This architecture is---unlike the one in \cref{eq:rebar-c}---well-defined for all inputs.
A drawback of using such control variate is that it can start out not being very correlated with $\log(\frac{1}{K} \sum_{k = 1}^K w_k)$.

\Acrshort{RELAX} also proposes using a summation of the free-form control variate like the one in \cref{eq:relax-c} and a more correlated control variate in \cref{eq:rebar-c}.

We have tried all architectures and found that \cref{eq:relax-c} leads to the most stable and best training.

Using \acrshort{REBAR}/\acrshort{RELAX} for more complicated models is possible, however designing an architecture that is highly correlated with the high-variance term and stable to train still remains a challenge.

\begin{figure*}[!htb]
  \centering
  \includegraphics[width=\textwidth]{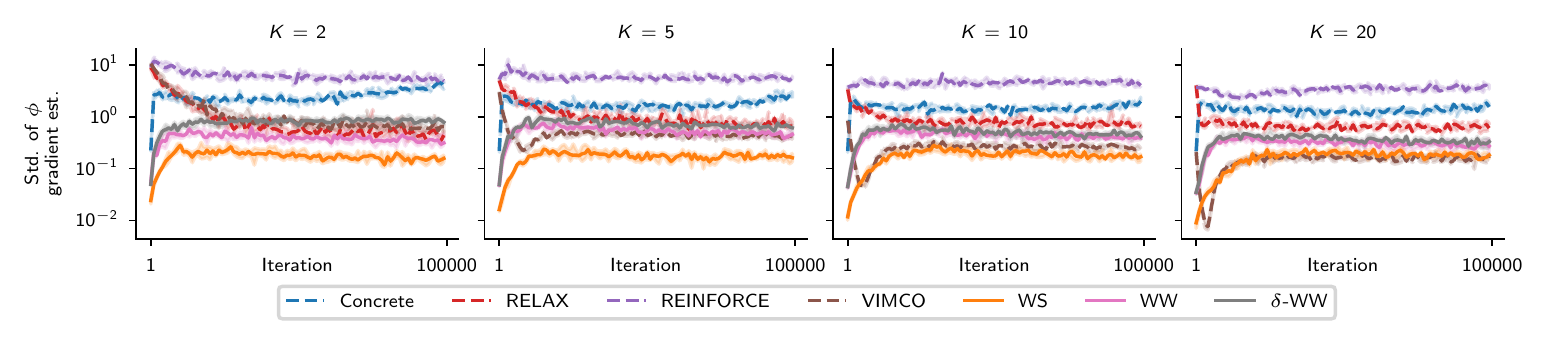}
  \vspace*{-4ex}
  \caption{
    Standard deviation of gradient estimator of $\phi$ for \gls{GMM}.
    Median and interquartile ranges from $10$ repeats shown.
    \Gls{WW} and \gls{WS} have lower-variance gradient estimators of $\phi$ than \gls{IWAE} except \gls{VIMCO}, as they avoid the high-variance term \circled{1} in \eqref{eq:iwae-reinforce}.
    This is a necessary, but not sufficient, condition for efficient learning, with other factors being gradient direction and the ability to escape local optima.
    The standard deviation of $\phi$'s gradient estimator is given by $\frac{1}{D_\phi} \sum_{d = 1}^{D_\phi} \std(g_d)$ where $g_d$ is the $d$th (out of $D_\phi$) element of one of $\phi$'s gradient estimators (e.g. \cref{eq:iwae-reinforce} for \acrshort{REINFORCE}) and $\std(\cdot)$ is estimated using $10$ samples.
  }
  \label{fig:gmm_just_std}
  \vspace*{-2ex}
\end{figure*}

\subsection{ADDITIONAL RESULTS}
\label{app:gmm/additional}

\begin{figure*}[!ht]
  \centering
  \includegraphics[width=\textwidth]{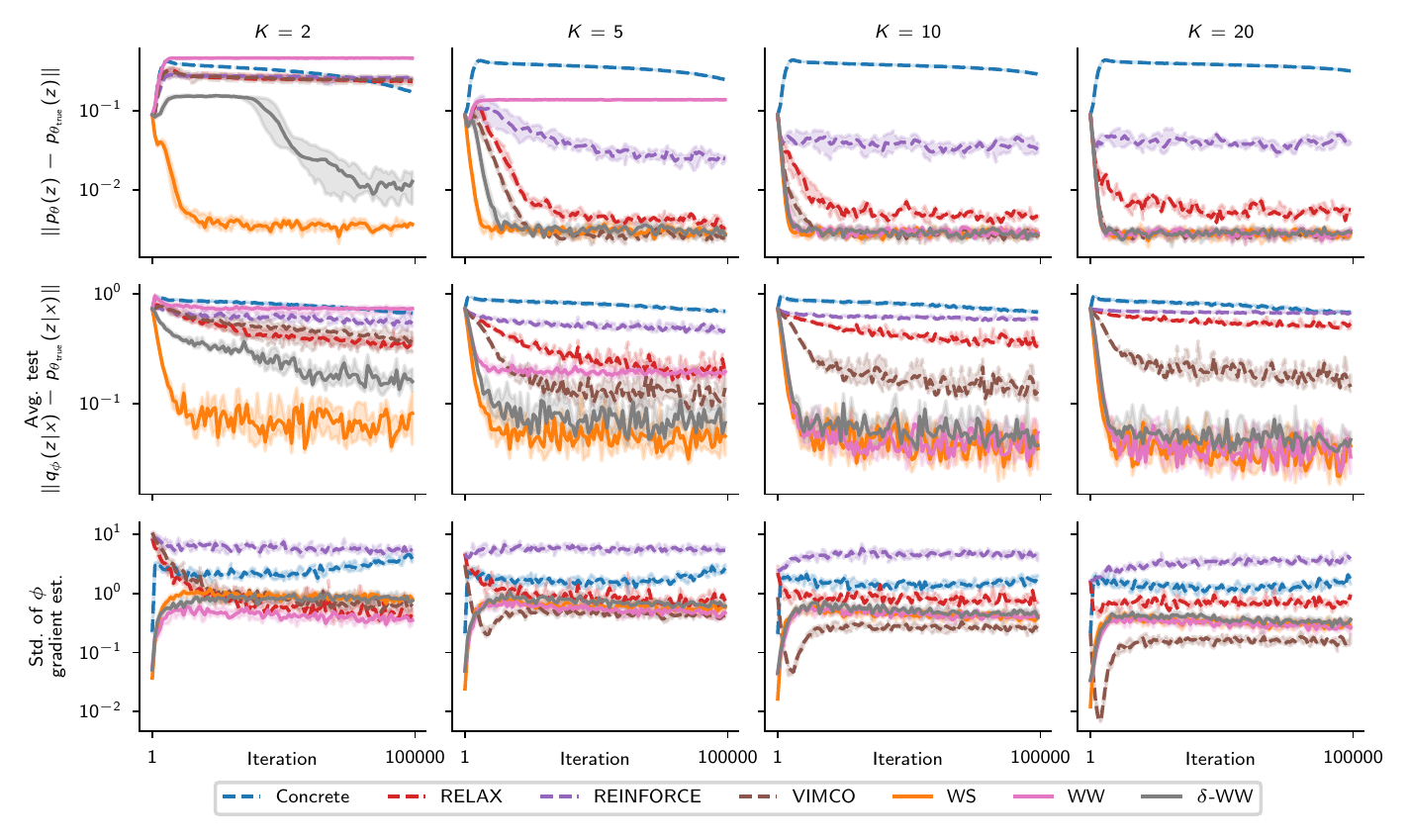}
  \vspace*{-4ex}
  \caption{
    \Gls{GMM} training when $p_\theta(x)$ is close to $p_{\theta^*}(x)$.
    \Gls{WS} outperforms other methods including \gls{WW} in generative model (top) and inference network (middle) learning.
    \Gls{VIMCO} has the lowest gradient variance (bottom) but still performs worse than \gls{WS} and results in worsening of the inference network as number of particles is increased.
  }
  \label{fig:gmm_init_near}
  \vspace*{-2ex}
\end{figure*}

Here, we include additional \gls{GMM} experiments:
one for studying $\phi$'s gradient variance (\cref{fig:gmm_just_std}),
the other for comparing performances of the generative model and inference networks when $\theta$ is initialized closer to $\theta^*$ than in the main paper (\cref{fig:gmm_init_near}).

\Gls{WW} and \gls{WS} have lower variance gradient estimators than \gls{IWAE}, except \gls{VIMCO}.
This is because $\phi$'s gradient estimators for \gls{WW} and \gls{WS} do not include the high-variance term \circled{1} in \cref{eq:iwae-reinforce}.
This is a necessary but not sufficient condition for efficient learning with other important factors being gradient direction and the ability to escape local optima.
Employing the Concrete distribution gives low-variance gradients for $\phi$ to begin with, but the model learns poorly due to the high gradients bias (due to high temperature hyperparameter).

In \cref{fig:gmm_init_near}, we initialize $\theta$ so that the mixture probabilities are constant.
This means that the data bias is smaller than in the main paper's setting.
With smaller data bias, we expect \gls{WS} to perform better.
This is empirically verified since \gls{WS} outperforms other methods, including \gls{WW}.

\section{SIGMOID BELIEF NETWORKS}
\label{app:sigmoid_belief_nets}

\begin{figure*}[!ht]
  \centering
  \includegraphics[width=0.99\textwidth]{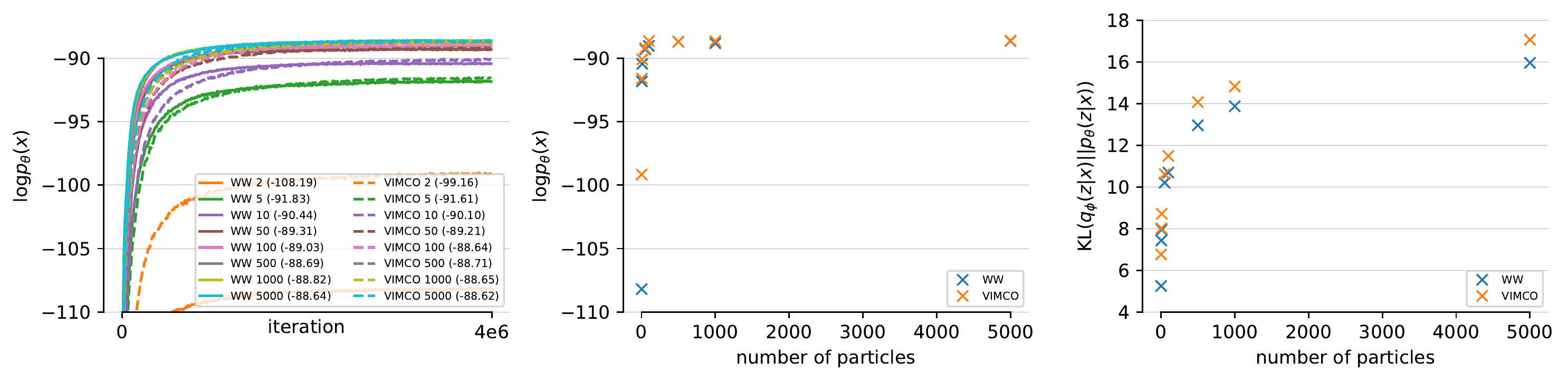}
  \vspace*{-4ex}
  \caption{
    Training of sigmoid belief nets.
    \emph{(Left)}
    Training curves:
    \gls{WW} learns faster than \gls{VIMCO} but results in equal or slightly worse end test log likelihood.
    \emph{(Middle)}
    Log evidence values at the end of training:
    \gls{VIMCO} is slightly better than \gls{WW} in low-particle regimes but virtually the same in high-particle regimes.
    \emph{(Right)}
    \gls{KL} divergence at the end of training:
    \gls{WW} results in much lower \gls{KL} divergence than \gls{VIMCO}.
  }
  \label{fig:sbn}
  \vspace*{-2ex}
\end{figure*}

In \cref{fig:sbn}, we show training of sigmoid belief networks with three stochastic layers with the same architecture as in \citet{mnih2016variational}.
We additionally drive number of particles up to $K = 5000$ and include \gls{KL} plots.
We find that in high particle regimes, model learning is virtually the same for \gls{WW} and \gls{VIMCO}.
However, \gls{WW} outperforms \gls{VIMCO} in terms of inference network learning.

\section{DISCRETE VAES}
\label{app:dvae}

Rolfe 2016 \citep{rolfe2016dvae} introduces discrete \textsc{vae} (\textsc{dvae}).
It combines a prior over binary latent variables with an element-wise spike-and-X smoothing transformation, allowing approximate marginalization of the discrete variables.
This results in a continuous relaxation of discrete variables and a low-variance gradient estimator.
\citet{vahdat2018dvaepp} replaced the original transformation with an overlapping exponential transformation, leading to a yet lower-variance gradient estimator.
While both approaches produce relaxed binary variables, the relaxation is  significantly less tight (\cite{vahdat2018dvaepp}, Appendix C, Figure 5.) then the \textsc{concrete} of \cite{jang2017categorical,maddison2017concrete}
Both approaches require analytical inverse \textsc{cdf}s of the smoothing transformations, a shortcoming addressed by \citet{vahdat2018dvaehash} --- it also leads to a tighter relaxation than its predecessors, however no comparison to \textsc{concrete} is available.

\textsc{Dvae} was designed for undirected binary priors, \textit{e.g.}\ restricted Boltzmann machines (\textsc{rbm}), and it does not account for the case of categorical latent variables.
It is possible to construct a $d$-dimensional categorical variable from  $d-1$ binary variables via stick-breaking construction.
This process is slow, however, as it requires $\operatorname{\mathcal{O}}(d)$ sequential operations and cannot be parallelized.
Moreover, in the case of relaxed variables, the tightness of the derived relaxed categorical variable decreases exponentially with the number of dimensions.
This is a major issue in control flows: not only we have to evaluate all branches of the control flow, but the indicator variables that we multiply with outcomes of different branches become exponentially loose with the depth of the flow.

\end{document}